\documentclass[lettersize,journal]{IEEEtran}
\usepackage{amsmath,amsfonts}
\usepackage{algorithmic}
\usepackage{algorithm}
\usepackage{array}
\usepackage[caption=false,font=normalsize,labelfont=sf,textfont=sf]{subfig}
\usepackage{textcomp}
\usepackage{stfloats}
\usepackage{url}
\usepackage{verbatim}
\usepackage{graphicx}
\usepackage{cite}
\usepackage{booktabs} % For formal tables
\usepackage{xcolor}
\usepackage{soul}
\usepackage{epsfig}
\usepackage{ulem}
\usepackage{amssymb}
\usepackage{helvet}
\usepackage{courier}
\usepackage{multirow}
\usepackage{paralist}
\usepackage{balance}
\usepackage{utfsym}    % colors
\usepackage[numbers]{natbib}

\newcommand{\PreserveBackslash}[1]{\let\temp=\\#1\let\\=\temp}
\newcolumntype{C}[1]{>{\PreserveBackslash\centering}p{#1}}
\newcolumntype{R}[1]{>{\PreserveBackslash\raggedleft}p{#1}}
\newcolumntype{L}[1]{>{\PreserveBackslash\raggedright}p{#1}}

\newcommand{\p}{{\bf p}}

\newcommand{\y}{{\bf y}}

\newcommand{\E}{\mathbb{E}}
\newcommand{\R}{\mathbb{R}}

\begin{document}

\title{Rebalanced Vision-Language Retrieval Considering Structure-Aware Distillation}

\author{
Yang Yang,~\IEEEmembership{Member,~IEEE,}
Wenjuan Xi,~\IEEEmembership{Student Member,~IEEE,}
Luping Zhou,~\IEEEmembership{Senior Member,~IEEE,}\\
Jinhui Tang,~\IEEEmembership{Senior Member,~IEEE}
\thanks{Manuscript received March 16, 2024; revised September 30, 2024 and November 12, 2024; accepted December 3, 2024. Date of current version December 12, 2024.
This work was supported 
% in part by the National Key RD Program of China under Grant 2022YFF0712100, 
% in part by the NSFC under Grant 62276131, 
% in part by the Natural Science Foundation of Jiangsu Province of China under Grant BK20240081 and Grant BG2024042, 
% and in part by the Fundamental Research Funds for the Central Universities under Grant No.30922010317.
in part by the National Key RD Program of China (2022YFF0712100), 
in part by the NSFC (62276131), 
in part by the Natural Science Foundation of Jiangsu Province of China under Grant (BK20240081, BG2024042), 
and in part by the Fundamental Research Funds for the Central Universities (No.30922010317).
The associate editor coordinating the review of this manuscript and approving it for publication was Dr. Sicheng Zhao. \textit{(Corresponding author: Jinhui Tang.)}
}
\thanks{Yang Yang, Wenjuan Xi and Jinhui Tang are with the School of Computer Science and Engineering, Nanjing University of Science and Technology, Nanjing 210094, China (e-mail: yyang@njust.edu.cn; xiwenjuan@njust.edu.cn; jinhuitang@njust.edu.cn). Luping Zhou is with the School of Electrical and Information Engineering, The University of Sydney, Sydney, NSW 2006, Australia (e-mail: luping.zhou@sydney.edu.au).}
\thanks{This paper has supplementary downloadable material available at http://ieeexplore.ieee.org., provided by the author. The material includes supplemental material. Contact yyang@njust.edu.cn for further questions about this work.}
}

\markboth{IEEE TRANSACTIONS ON IMAGE PROCESSING, December~2024}%
{Shell \MakeLowercase{\textit{et al.}}: A Sample Article Using IEEEtran.cls for IEEE Journals}



\maketitle

\begin{abstract}
Vision-language retrieval aims to search for similar instances in one modality based on queries from another modality. The primary objective is to learn cross-modal matching representations in a latent common space. Actually, the assumption underlying cross-modal matching is modal balance, where each modality contains sufficient information to represent the others. However, noise interference and modality insufficiency often lead to modal imbalance, making it a common phenomenon in practice. The impact of imbalance on retrieval performance remains an open question. In this paper, we first demonstrate that ultimate cross-modal matching is generally sub-optimal for cross-modal retrieval when imbalanced modalities exist. The structure of instances in the common space is inherently influenced when facing imbalanced modalities, posing a challenge to cross-modal similarity measurement. To address this issue, we emphasize the importance of meaningful structure-preserved matching. Accordingly, we propose a simple yet effective method to rebalance cross-modal matching by learning structure-preserved matching representations. Specifically, we design a novel multi-granularity cross-modal matching that incorporates structure-aware distillation alongside the cross-modal matching loss. While the cross-modal matching loss constraints instance-level matching, the structure-aware distillation further regularizes the geometric consistency between learned matching representations and intra-modal representations through the developed relational matching. Extensive experiments on different datasets affirm the superior cross-modal retrieval performance of our approach, simultaneously enhancing single-modal retrieval capabilities compared to the baseline models.
\end{abstract}


\section{Introduction}
\IEEEPARstart{I}{n} re-examining current vision-language retrieval tasks~\cite{LuBPL19,DBLP:journals/ipm/HuangLZLWJQ20,DCD:journals/tmm/RaoDQFLST23}, approaches typically focus on searching images for given texts or retrieving texts from image queries~\cite{wan2024covlr}. Unlike single-modal retrieval, the primary challenge in vision-language retrieval lies in the semantic divergence of heterogeneous data, necessitating effective constraints to ensure the consistency between two modal representations~\cite{DBLP:journals/tip/ZhangZWTL21}.

To solve this problem, state-of-the-art~(SOTA) approaches~\cite{LeeCHHH18, DBLP:journals/pami/HuHPWP23} usually adopt various modal interaction modules to govern the matching of aligned modal representations for each instance, i.e., pulling cross-modal representations closer while distancing them from other instances. Initial methods are dual-encoder approaches~\cite{FaghriFKF18, LeeCHHH18, DiaoZML21, LuBPL19}, in which the two modalities interact at the \textbf{output level}. These methods typically build independent embedding networks for each modality, and then design coarse-grained (e.g., global-level~\cite{FaghriFKF18}) or fine-grained (e.g., region-level~\cite{LeeCHHH18, DBLP:journals/tip/PengQY18, DBLP:conf/aaai/ZhangMZ022} and graph-level~\cite{Lu2021}) similarity functions to measure the matching degree of cross-modal output representations. With the development of transformer-based architectures, vision-language transformers are proposed~\cite{DBLP:journals/tip/LuoZSWCWHJ22}, in which the two modalities interact from the \textbf{input level}. These methods adopt the deep transformer as a modal interaction module to collectively model the concatenation of two modal inputs, which can allow cross-modal instances to represent each other effectively.

\begin{figure}[t] 
\begin{minipage}[t]{0.48\linewidth}
\includegraphics[width=\linewidth]{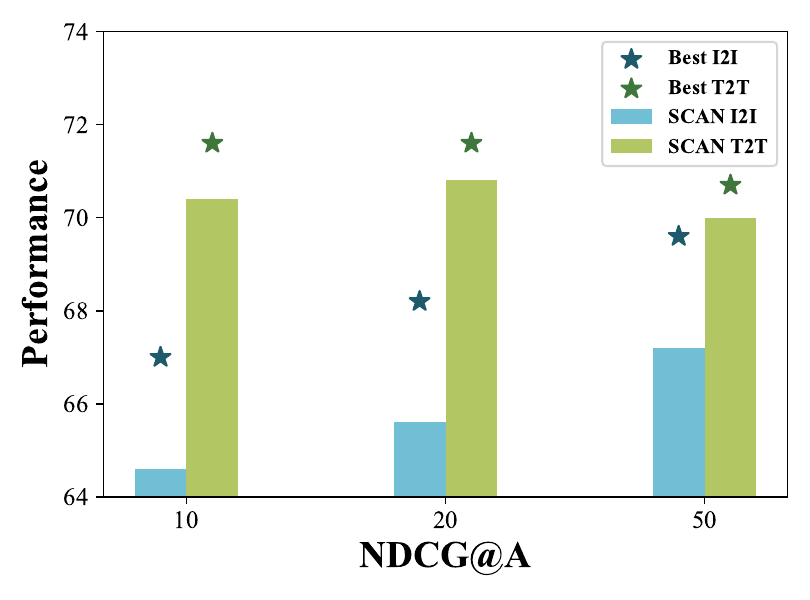}\\
{\small (a). Both modalities are simultaneously adversely affected.}
\end{minipage} 
\begin{minipage}[t]{0.48\linewidth}
\includegraphics[width=\linewidth]{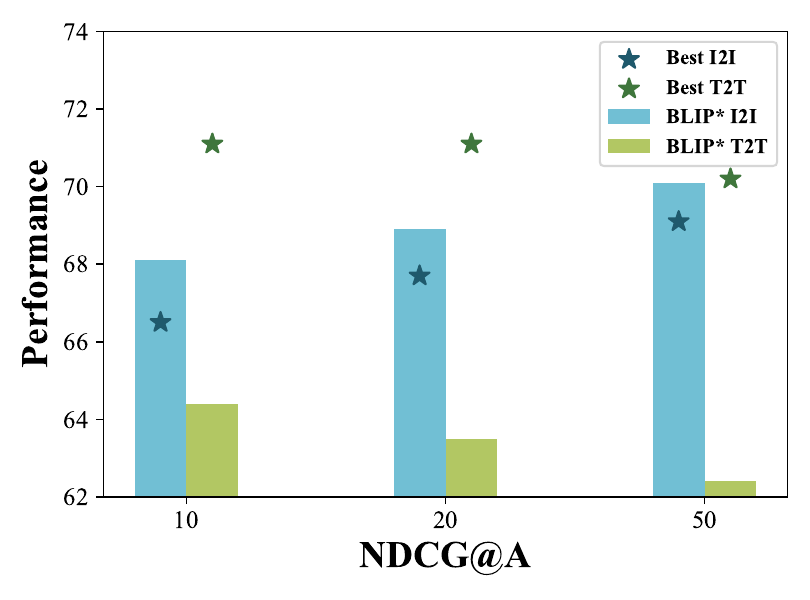}\\
{\small (b). ``Weak'' modality slightly improves, ``strong'' modality seriously declined.}
\end{minipage}
\caption{The impact of imbalanced modalities on single-modal retrieval in cross-modal learning. Figures (a) and (b) respectively present the NDCG@\{10,20,50\} results of the single-modal encoders after training two cross-modal models, SCAN ~\cite{LeeCHHH18} and BLIP ~\cite{DBLP:conf/icml/0001LXH22} on the MS-COCO (1K) dataset. The asterisk (*) indicates that the large models were retrained from scratch. ``Best'' denotes the results obtained by the best single-modal models trained separately on images and texts. The cross-modal model and single-modal models adopt identical network architectures.}\label{fig:imbalanced}
\end{figure}

\begin{figure*}[t]
\centering
\includegraphics[width=150mm]{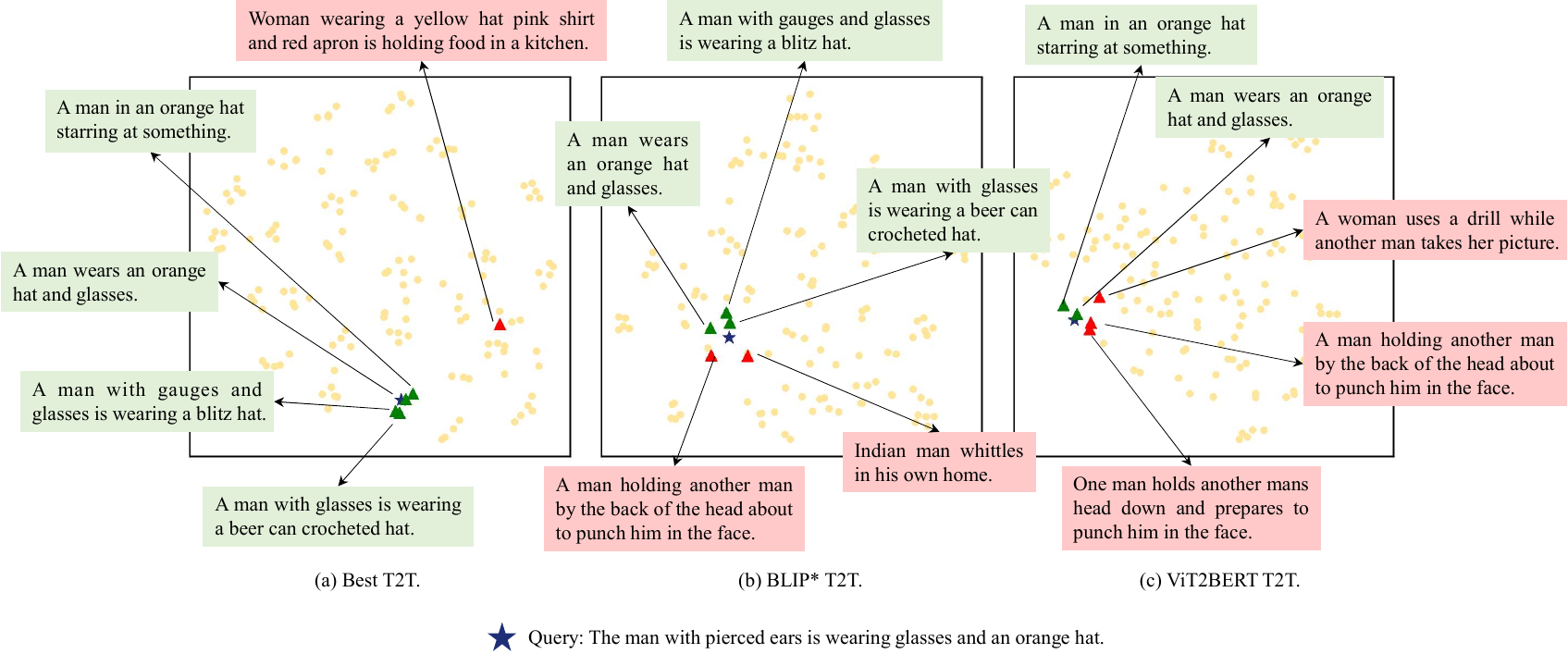}
\caption{T-SNE visualization of Best T2T, BLIP* T2T, ViT2BERT T2T on the FLICKR30K dataset, where BLIP* T2T uses the ViT/B and BERT as backbones. We randomly choose a text query and a database with 150 samples. The blue pentagram represents the text query, while the top-5 retrieval candidates are shown as triangles. Ground-truth candidates are marked in green, and non-ground-truth candidates are marked in red.}\label{fig:tsne}
\end{figure*}

Cross-modal matching representation learning~\cite{DBLP:conf/cvpr/FuMSZ23} can be regarded as a variant of unsupervised single-modal contrastive learning, with the aligned modal instances serving as the positive anchors. However, experiments suggest that the effectiveness of aligned modal instances is inferior to the data augmentation technique in single-modal contrastive learning. This discrepancy arises from the noise interference and modality insufficiency in real applications~\cite{WangZ13,10132374}, leading to notorious modality imbalance phenomenon~\cite{YangYZJ15, ModalityCompetition:conf/icml/HuangLZYH22}. Here, the modality with greater sufficiency is defined as strong modality~\cite{ModalityCompetition:conf/icml/HuangLZYH22}, while the less sufficient one is termed weak modality. In real-world applications, we can employ the performance of the optimal single-modal model to assess the “strength” of a modality~\cite{WangZ13}. We use an experiment for illustration in Fig.~\ref{fig:imbalanced}. From Fig.~\ref{fig:imbalanced}, we observe that the text-to-text~(T2T) retrieval performance of the text model BERT~\cite{DevlinCLT19} consistently outperformed the image-to-image~(I2I) retrieval performance of the vision model Swin Transformer~(SwinT)~\cite{LiuL00W0LG21}, designating text as the ``strong'' modality and image as the ``weak'' modality. Furthermore, a simultaneous decrease in the expressive capabilities of both the weak and strong modalities, as shown in Fig.~\ref{fig:imbalanced}(a), and a slight improvement in the performance of the weak modality, accompanied by a severe decline in the performance of the strong modality, as shown in Fig.~\ref{fig:imbalanced}(b). This suggests that enforcing cross-modal consistency learning disrupts the optimal instance structure within single-modal spaces, thereby diminishing the accuracy of similarity-based retrieval. This issue is further evident in the t-SNE~\cite{van2008visualizing} visualization in Fig.~\ref{fig:tsne}, where the structural information in the text of BLIP* and ViT2Bert is partially compromised compared with the best text retrieval model BERT.

To address the modal imbalance problem, coined as the information gap between the two modalities, we propose a shift from exact instance-level modal matching to the structure preservation of the learned representations in cross-modal representation learning. In particular, we design a multi-granularity distillation module to rebalance cross-modal matching, which improves instance-level matching through representation-level and structure-aware distillation to construct better latent structures. The newly introduced geometric consistency can promote geometric symmetry in the latent common space. Specifically, for inter- and intra-modal structure preservation, we introduce two modal-independent teacher networks, which can jointly distill the cross-modal model using the designed relational matching. Moreover, to obtain the optimal instances' relationships from teacher models, we adaptively learn the fusion coefficient of two teachers. In summary, the main contributions of this paper are as follows: 1). We analyze the impact of the exact matching on cross-modal retrieval tasks. We show that reducing the modal gap at the instance-level does not always ensure better latent structure particularly when dealing with imbalanced modalities. Instead, a large information gap between the modalities can hurt the performance. 2). We advocate preserving both semantic and structural consistency, and propose the multi-granularity distillation on top of the cross-modal consistency loss, enhancing multi-granularity matching in latent common space. 3). Through extensive experiments, our approach shows improved performance across cross-modal, single-modal, and mixed retrieval. This validates the effectiveness of incorporating relational knowledge in the learning of comprehensive matching representations.

\begin{table*}[!htb]{\footnotesize
\centering
\caption{Single-modal retrieval performance comparison. Evaluation criteria is NDCG$@$A ($@$A for simplicity).}
\label{tab:tab}
\begin{tabular*}{0.985\textwidth}{@{\extracolsep{\fill}}@{}l@{}|@{}c@{}|@{}c@{}|@{}c@{}|@{}c@{}|@{}c@{}|@{}c@{}|@{}c@{}|@{}c@{}|@{}c@{}|@{}c@{}|@{}c@{}|@{}c@{}|@{}c@{}|@{}c@{}|@{}c@{}|@{}c@{}|@{}c@{}|@{}c@{}|@{}c@{}|@{}c@{}|@{}c@{}|@{}c@{}|@{}c@{}|@{}c@{}}
\hline
\multirow{3}{*}{Methods} & \multicolumn{6}{c|}{MS-COCO (1K)} & \multicolumn{6}{c|}{MS-COCO (5K)} & \multicolumn{6}{c|}{FLICKR30K}& \multicolumn{6}{c}{Vizwiz}\\
\cline{2-25}
& \multicolumn{3}{c|}{ I2I } & \multicolumn{3}{c|}{ T2T } 	& \multicolumn{3}{c|}{I2I} & \multicolumn{3}{c|}{T2T} & \multicolumn{3}{c|}{I2I} & \multicolumn{3}{c|}{T2T} & \multicolumn{3}{c|}{I2I} & \multicolumn{3}{c}{T2T}\\
\cline{2-25}
& $@$10 & $@$20 & $@$50 & $@$10 & $@$20 & $@$50 & $@$10 & $@$20 & $@$50 & $@$10 & $@$20 & $@$50 & $@$10 & $@$20 & $@$50 & $@$10 & $@$20 & $@$50 & $@$10 & $@$20 & $@$50 & $@$10 & $@$20 & $@$50\\
\hline
ResNet101  &51.9 &54.3 &58.4 &- &- &- &42.9 &44.9 &48.3 &- &- &- &49.0 &51.6 &56.2 &- &- &- &49.4 &52.1 &56.5 &- &- &-\\
SwinT &66.5 &67.7 &69.1 &- &- &- &61.5 &63.2 &65.7 &- &- &- &64.1 &64.5 &67.9 &- &- &- &50.3 &53.9 &57.3 &- &- &-\\	
LSTM &- &- &- &63.6 &65.5 &66.9 &- &- &- &58.9 &61.2 &63.7 &- &- &- &57.9 &60.4 &62.1 &- &- &- &51.2 &54.0 &57.4\\
BERT &- &- &- &71.1 &71.1 &70.2 &- &- &- &63.4 &64.8 &66.1 &- &- &- &71.6 &70.2 &68.4 &- &- &- &55.5 &57.5 &59.6\\
\hline
\end{tabular*}}
\end{table*}

\begin{table}[t]{\small
\centering
\caption{Comparison with single-modal retrieval models~(SMR), the distillation image model taught by text model~(S2W@Image), and the distillation text model taught by image model~(W2S@Text).}
\label{tab:weak}
\begin{tabular*}{0.475\textwidth}{@{\extracolsep{\fill}}c|c|c|c|c|c|c}
\hline
\multirow{3}{*}{Method} & \multicolumn{3}{c|}{I2I}& \multicolumn{3}{c}{T2T}\\
\cline{2-7}
& $@$10 & $@$20 & $@$50 & $@$10 & $@$20 & $@$50  \\
\hline
SMR  &64.1 &64.5 &67.9 &\bf 71.6 & \bf 70.2 &\bf 68.4\\	
S2W@Image &\textbf{64.2} &\textbf{65.6} &\textbf{68.3} & - & - & -\\
W2S@Text  & - & - & - & {71.5} &{70.0} &{68.0}\\
\hline
\end{tabular*}
}
\end{table}

\section{Related Work}
\subsection{Vision-Language Retrieval}
To learn matching representations of heterogeneous modalities, a large number of vision-language retrieval models are proposed~\cite{DBLP:conf/ijcai/YangZXYZY21,DBLP:journals/tmm/YuZLQHTW20,DBLP:journals/pami/XuLYHS22, DBLP:journals/tcsv/DongLZNZ22}.
Traditional approaches~\cite{FaghriFKF18, LeeCHHH18, LuBPL19, DiaoZML21,MessinaAEFGM21} always utilize dual architecture, where the image and text modalities are separately embedded into a common space and then maximize the cross-modal representation similarity. For example, \cite{FaghriFKF18} built two independent modal encoders (i.e., VGG19 for image and GRU for text), and incorporated hard negatives in the triplet loss function; \cite{LeeCHHH18} further utilized the Faster R-CNN for image modality, and discovered the full latent alignments using both image regions and words in a sentence as context; 
CLIP~\cite{DBLP:conf/icml/RadfordKHRGASAM21} employed a large-scale contrastive pre-training approach to effectively align vision and language modalities; \cite{DBLP:journals/pami/HuZLPZP23} applied contrastive learning to cross-modal hashing using a momentum-based optimizer and a cross-modal ranking learning loss. These approaches can efficiently benefit from numerous simple modal interactions such as dot products or shallow attention layers~\cite{DBLP:journals/tip/PengQY18}. In contrast, inspired by the significant advances in language understanding led by Transformers~\cite{DevlinCLT19}, SOTA retrieval models turn to employ large vision-language transformers for modal interaction~\cite{TanB19, LuBPL19}. For example, \cite{TanB19} developed a large-scale cross-modal encoder with five diverse representative pre-training tasks; \cite{LuBPL19} processed both visual and textual inputs that interact through co-attention transformer layers; \cite{Li2021} introduced a contrastive loss to align the image and text representations before fusing through cross-modal attention. In these approaches, vision and language inputs are fed into a unified or separate cross-modal attention branch to compute the similarity between them, thereby obtaining more similar cross-modal representations. However, we experimentally observe that the learned matching representations from cross-modal retrieval approaches will be affected when existing imbalanced modalities.

\subsection{Imbalanced Multi-Modal Learning}
The important assumption behind cross-modal methods is the modal balance, i.e., it typically assumes that each modality can well represent other modalities~\cite{10132374}. 
However, in real-world scenarios, factors such as data noise multi-modal data~\cite{WangZ13}, heterogeneity of multi-modal data and missing multi-modal data can lead to modality imbalance. For instance, the noise can impact the model's ability to learn effective information, resulting in the sufficiency of different modalities is various, leading to the existence of strong and weak modalities~\cite{YangYZJ15, ModalityCompetition:conf/icml/HuangLZYH22}. In traditional cross-modal matching constraints, strong and weak modalities will interact with each other, and it is difficult to control the strong modal structure from being affected by the weak modality when exiting a large cross-modal divergence. This problem leads to multi-modal joint training being hard and has been researched in multi-modal fusion for classification tasks, \cite{wang2020makes, HanZFZ21, PengWD0H22} observed that the best single-modal network often outperforms the multi-modal networks. Therefore, \cite{wang2020makes} proposed to estimate the single-modal generalization and overfitting speeds to calibrate the learning through loss re-weighing; \cite{HanZFZ21} promoted reliability and robustness by integrating evidence that explained the prediction of each modality; \cite{PengWD0H22} chose to slow down the learning rate of the mighty modality by online modulation to lessen the inhibitory effect on the other modality. Nevertheless, these methods only focus on multi-modal fusion tasks by learning complementary information, which is different from the cross-modal retrieval for learning matching representations.

\begin{figure*}[t]
\centering
\includegraphics[width=160mm]{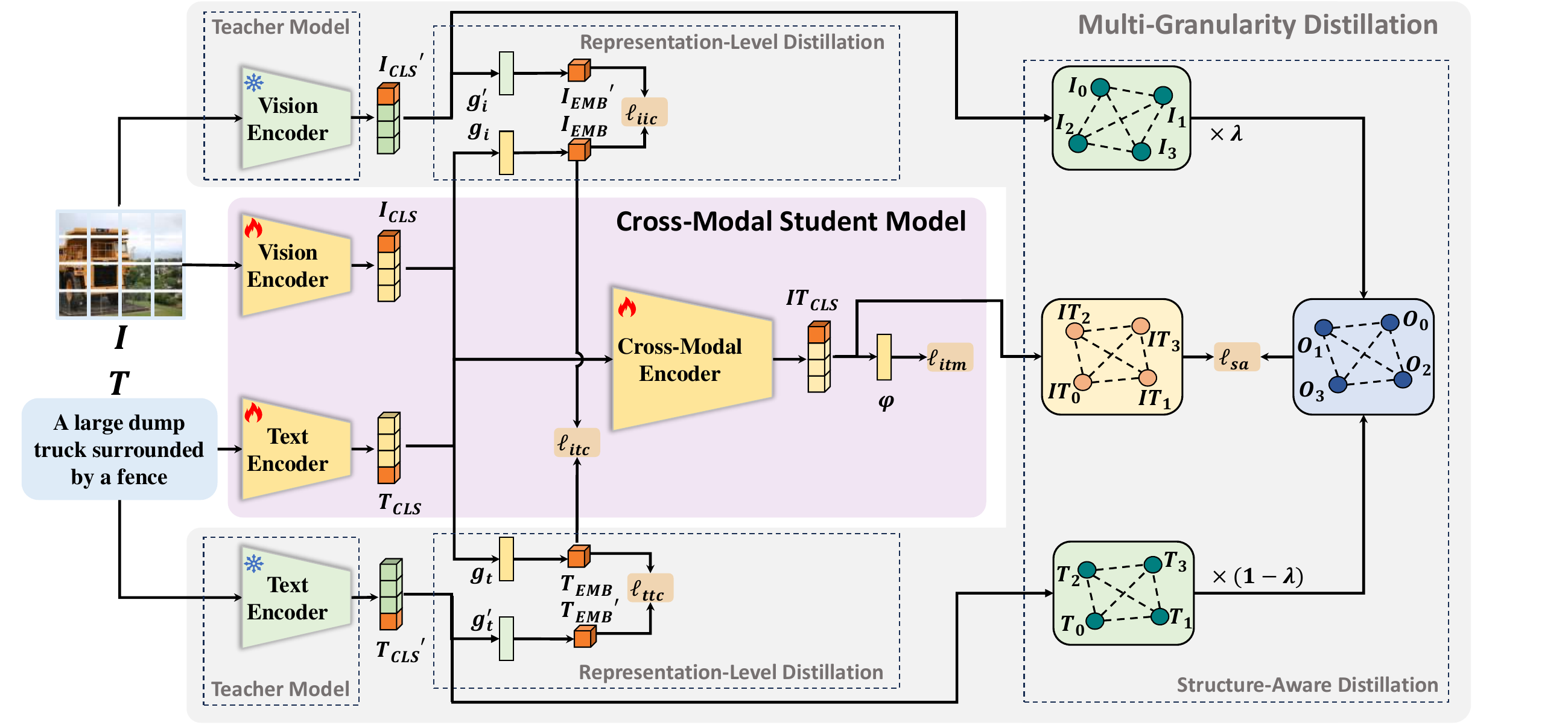}
\caption{Illustration of our framework. Expanding on the framework of cross-modal matching, we incorporate a single-modal teacher network. Our multi-granularity distillation includes representation-level distillation and structure-aware distillation, where the former optimizes the expressive capabilities of individual modalities via contrastive loss, and the latter enhances instance-level matching through structure-aware distillation.} \label{fig:fra}
\end{figure*}

\section{Proposed Method}
To rebalance the vision-language retrieval, we aim to transfer the optimal instances' structure from teacher networks to cross-modal model in learning matching representations. Therefore, the overall architecture consists of two networks: 1). a cross-modal student network, and 2). two-modal independent teacher networks. In this paper, we concentrate on the image and text modalities, considering the effectiveness and feasibility, we adopt the cross-modal network following~\cite{Hang2021} and utilize the Swin Transformer~\cite{LiuL00W0LG21}/BERT~\cite{DevlinCLT19} for image/text modalities. Notably, our method can be used as a plug-and-play module, so we can transform any state-of-the-art cross-modal network and independent networks in our framework, and more analyses are provided in the experiments part. In the following subsections, we will first provide the exploration of modal imbalance, then introduce our framework and learning objectives.
\subsection{Exploration of Modal Imbalance}
We investigate the phenomenon of imbalance between image and text modalities, specifically focusing on differences in modality sufficiency. Proposition 2 in \cite{WangZ13} indicates that modality sufficiency is correlated with the performance of the optimal model: greater insufficiency leads to poorer performance in single-modal models. Hence, we assess modality sufficiency based on single-modal retrieval performance. Specifically, we use the ResNet101~\cite{HeZRS16} and Swin Transformer~(SwinT)~\cite{LiuL00W0LG21} for training the image modality, and LSTM~\cite{HochreiterS97} and BERT~\cite{DevlinCLT19} for the text modality. The results are shown in Table~\ref{tab:tab}, which shows that text consistently outperforms images in single-modal retrieval across all datasets, indicating that the sufficiency of the text modality is greater than that of the image modality. Consequently, we categorize text as the strong modality and image as the weak modality. Furthermore, Swin Transformer and BERT exhibit the best retrieval performance for images and text, respectively.

Additionally, to further study whether modals can sufficiently represent each other, we conducted experiments where the strong modality text distills the weak modality image~(denoted as S2W@Image), and the weak modality distills the strong modality~(denoted as W2T@Text). Specifically, in the S2W@Image setting, BERT serves as the teacher model and SWIN as the student model, with vision-language contrastive learning applied between them.
The results are presented in Table~\ref{tab:weak}. We can find that: 1). The I2I performance of single-modal retrieval~(SMR) is worse than that of S2W@Image, demonstrating that the strong modality, provides sufficient information to represent the weak modality; 2). The T2T performance of SMR is better than that of W2S@Text, demonstrating that the weak modality image cannot provide sufficient information to represent the strong modality text.

\subsection{Cross-Modal Student Model}
We use X-VLM~\cite{Hang2021} without the bounding box module as the cross-modal student model.

\noindent {\bf{Vision and Language Encoders:}} Vision encoder produces fine-grained visual concept representations based on the SOTA Swin Transformer. As shown in Fig.~\ref{fig:fra}, Swin Transformer splits an image $I \in \R^{C \times H \times W}$ into patches and flats to $I \in \R^{(C \times P^2) \times L_I}$, where $P \times P$ represents the patch resolution and $L_I = HW/P^2$ represents the number of patches following~\cite{DosovitskiyB0WZ21}. On the other hand, the language encoder maps the input text $T \in \R^{L_T \times d}$ to the same dimension subspace of the image with BERT, $L_T$ is the words number, and $d$ is the dimension of the common subspace. We recommend referring to \cite{LiuL00W0LG21, DosovitskiyB0WZ21} for more detailed information.

\noindent {\bf{Cross-Modal Encoder:}} It refers to the cross-attention transformer, i.e., the text-oriented cross-attention transformer, which aims to harness the efficacy of interaction layers to process visual and textual representations. As shown in Fig.~\ref{fig:fra}, the cross-attention can be formulated as:
\begin{equation}\label{eq:e0}
\begin{split}
Att(T)  & = softmax(\frac{Q_T K_I^\top}{\sqrt{d_M}})V_I, \\
\end{split}
\end{equation} 
where $Q_{T} = T W_{qT}$, $K_{I} = I W_{kI}$, and $V_{I} = I W_{vI}$ represent queries, keys, and values. $W_{qT} \in \R^{d_T \times d_M}, W_{kI} \in \R^{d_I \times d_M}$, and $W_{vI} \in \R^{d_I \times d_M}$ are learnable mapping matrixes. Besides, multi-head attention is composed of $M$ parallel heads, and $d_M = d/M$. As a result, the other modal representations (i.e., query) are fused with the self-representations (i.e., keys and values) at each attention layer $Att(T)$ in Equation \ref{eq:e0}. Note that we directly adopt the pre-trained network~\cite{Hang2021} for fine-tuning.

\noindent {\bf{Cross-Modal Matching:}} Cross-modal matching aims to learn consistent cross-modal representations, including vision-language contrastive learning (ITC) and vision-language matching (ITM). Following X-VLM~\cite{Hang2021}, ITC preliminarily learns cross-modal consistency representations by treating all samples equally and performing contrastive learning across every pair of samples within the batch, thereby facilitating the initial filtering of similar sample pairs prior to vision-language fusion. In detail, we randomly sample a mini-batch of $J$ pairs and calculate the in-batch image-to-text and text-to-image similarity. This process can be represented as:
\begin{small}
\begin{align}
\ell_{itc} = \frac{1}{2}\E_{(I,T)}[CE(&\y^{i2t}(I),\p^{i2t}(I)) + CE(\y^{t2i}(T),\p^{t2i}(T))], \nonumber\\
p_k^{i2t}(I) &= \frac{exp(d(I,T_k)/\tau)}{\sum_{j=1}^J exp(d(I,T_j)/\tau)}, \label{eq:e2}\\ 
p_k^{t2i}(T) &= \frac{exp(d(T,I_k)/\tau)}{\sum_{j=1}^J exp(d(T,I_j)/\tau)}.\nonumber
\end{align}
\end{small}
Here, $\y^{i2t}(I),\y^{t2i}(T) \in \{0, 1\}^J$ denote the matching ground-truth, where $y_j^{i2t} = 1$ if $(I,T_j)$ is matched and $y_j^{i2t} = 0$ otherwise. $\p^{i2t}(I)$ and $\p^{t2i}(T)$ represent the in-batch image-to-text and text-to-image similarity, respectively. $d(I, T) = \cos(g_{i}(I_{CLS}), g_{t}(T_{CLS}))$ and $d(T, I) = \cos(g_{t}(T_{CLS}), g_{i}(I_{CLS}))$ denote the similarity, $I_{CLS}$ and $T_{CLS}$ are the [CLS] embeddings output by the visual and language encoders, respectively. $g_{t}$ and $g_{i}$ are linear transformations that map the [CLS] embeddings to a lower dimension. The instances in the same batch act as the negative anchor. $\tau$ denotes the temperature scale parameter. $CE$ is the cross-entropy loss. 

Consistent with the X-VLM~\cite{Hang2021}, ITM leverages hard negative sampling within the batch to predict whether a given image-text pair is a match, thereby enabling the learning of more refined margins. In detail, for each image in a mini-batch, we sample a hard negative text according to $p_k^{t2i}(T)$ in Equation \ref{eq:e2}. We use the [CLS] embedding output $IT_{CLS}$ of the cross-modal encoder to predict the matching probability, and the loss can be represented as:
\begin{equation}
\begin{split}
\ell_{itm} & = \E_{(I,T)}[CE(\y^{match},\p^{match})],\nonumber
\end{split}
\end{equation}
where $\y^{match}$ is a 2-dimensional one-hot vector representing the ground-truth label, $\p^{match} = \phi(IT_{CLS})$ represents the matching prediction output by the binary classifier $\phi(\cdot)$.  Therefore, the overall cross-modal loss can be formulated as: $\ell_{cr} = \ell_{itc} + \ell_{itm}$.
\subsection{Multi-Granularity Distillation}\label{sec:s3}
We employ single-modal teacher models to guide the cross-modal model, implementing multi-granularity distillation that incorporates structural distillation~\cite{ParkKLC19} on top of representational distillation. This approach preserves the structural consistency of cross-modal representations within the latent space, functioning as a rebalancing mechanism to enhance cross-modal retrieval performance.

\noindent {\bf{Single-Modal Teacher Models:}} We construct advantaged teacher models for two modalities, which aim to transfer structural knowledge for the cross-modal network in learning matching representations. Note that traditional single-modal contrastive learning usually pre-trains models pairwisely, leading to some semantically similar instances being regarded as antagonistic pairs. Therefore, inspired by~\cite{000300LYS21}, we employ unsupervised prototype-aware contrastive learning for preserving the single-modal structure, which can distinguish intra- and inter-cluster pairs with a distance regularization. 
% Given that the label might be available in practice, we also study the impact of the teacher model with supervised learning in Section~\uppercase\expandafter{\romannumeral4} of supplementary.
Given that labels may be available in practical scenarios, we investigate the impact of the teacher model under supervised learning. Furthermore, we examine the performance of pre-trained models, DINO~\cite{DBLP:conf/iclr/0097LL000NS23} and T5~\cite{DBLP:journals/jmlr/RaffelSRLNMZLL20}. The results of both experiments are provided in Section~\uppercase\expandafter{\romannumeral4} of the supplementary material.

\noindent {\bf{Representation-Level Distillation:}} At the single-modal representation level, we employ contrastive learning to narrow the expression gap between the student model and the teacher model, thereby mitigating the influence on single-modal representations during cross-modal matching. Specifically, with the randomly sampled batch pairs $J$, we calculate the similarity between the output of the visual model and the visual teacher model, as well as the output of the language model and the language teacher model. Each anchor instance's single-modal representations, paired with the corresponding teacher model output, form positive pairs, while the instances from the batch form negative pairs. The optimization objectives are as follows:
\begin{align}
&\ell_{iic} = \mathbb{E}_{(I,I)}[CE(\y^{i2i}(I),\p^{i2i}(I))], \nonumber\\
&\ell_{ttc} = \mathbb{E}_{(T,T)}[CE(\y^{t2t}(T),\p^{t2t}(T))], \label{eq:e10}\\
&p_k^{i2i}(I) = \frac{exp(d(I,I_k)/\tau)}{\sum_{j=1}^J exp(d(I,I_j)/\tau)}, \nonumber\\
&p_k^{t2t}(T) = \frac{exp(d(T,T_k)/\tau)}{\sum_{j=1}^J exp(d(T,T_j)/\tau)},\nonumber
\end{align}
where $\mathbf{y}^{i2i}(I), \mathbf{y}^{t2t}(T) \in \{0, 1\}^J$ denote the matching ground-truth, which are defined similarly to $\mathbf{y}^{i2t}(I), \mathbf{y}^{t2i}(T)$. $\p^{i2i}(I)$ denotes the in-batch image-to-image similarity between the outputs of student and teacher model. And $\p^{t2t}(T)$ is defined similarly. The $\ell_{iic}$ and $\ell_{ttc}$ aims to align the single-modal representations of each instance with their corresponding teacher model outputs. This ensures that the learning process does not disrupt the internal similarity relationships within the single-modal representations, thereby facilitating the effective transfer of structural knowledge to the cross-modal model.

\begin{table*}[htbp]{
\centering
\caption{Cross-modal retrieval performance comparison. Evaluation criteria is R@A. The method with “+" sign is our method.}
\label{tab:tab1}
\begin{tabular*}{0.985\textwidth}{@{\extracolsep{\fill}}@{}l@{\hspace{0.06cm}}|@{}c@{\hspace{0.06cm}}|@{}c@{\hspace{0.06cm}}|@{}c@{\hspace{0.06cm}}|@{}c@{\hspace{0.06cm}}|@{}c@{\hspace{0.06cm}}|@{}c@{\hspace{0.06cm}}|@{}c@{\hspace{0.06cm}}|@{}c@{\hspace{0.06cm}}|@{}c@{\hspace{0.06cm}}|@{}c@{\hspace{0.06cm}}|@{}c@{\hspace{0.06cm}}|@{}c@{\hspace{0.06cm}}|@{}c@{\hspace{0.06cm}}|@{}c@{\hspace{0.06cm}}|@{}c@{\hspace{0.06cm}}|@{}c@{\hspace{0.06cm}}|@{}c@{\hspace{0.06cm}}|@{}c@{\hspace{0.06cm}}|@{}c@{\hspace{0.06cm}}|@{}c@{\hspace{0.06cm}}|@{}c@{\hspace{0.06cm}}|@{}c@{\hspace{0.06cm}}|@{}c@{\hspace{0.06cm}}|@{}c@{\hspace{0.06cm}}}

\hline
\multirow{3}{*}{Methods} & \multicolumn{6}{c|}{MS-COCO (1K)} & \multicolumn{6}{c|}{MS-COCO (5K)} & \multicolumn{6}{c|}{FLICKR30K} & \multicolumn{6}{c}{Vizwiz}\\
\cline{2-25}
& \multicolumn{3}{c|}{I2T} & \multicolumn{3}{c|}{T2I} 	& \multicolumn{3}{c|}{I2T} & \multicolumn{3}{c|}{T2I} & \multicolumn{3}{c|}{I2T} & \multicolumn{3}{c|}{T2I} & \multicolumn{3}{c|}{I2T} & \multicolumn{3}{c}{T2I}\\
\cline{2-25}
& $@$1 & $@$5 & $@$10  & $@$1 & $@$5 & $@$10 & $@$1 & $@$5 & $@$10 & $@$1 & $@$5 & $@$10 & $@$1 & $@$5 & $@$10 & $@$1 & $@$5 & $@$10 & $@$1 & $@$5 & $@$10 & $@$1 & $@$5 & $@$10\\
\hline
VSE++ & 49.5 & 81.0 &90.0 &38.1 &73.3 &85.1 &39.0 &67.9 &79.5 &29.3 &59.1 &72.4 &31.6 &59.3 &71.7 &21.6 &50.7 &63.8 &35.1 &58.1 &65.4 &25.3 &48.1 &58.4\\
% \hline
SCAN  &72.7 &94.8 &98.4 &58.8 &88.4 &94.8 &48.3 &82.0 &89.1 &38.1 &68.9 &80.2 &60.0 &83.9 &90.7 &37.7 &66.3 &76.0 &34.9 &60.9 &72.8 &24.8 &48.6 &59.0\\
% \hline
IMRAM &76.7 &95.6 &98.5 &61.7 &89.1 &95.0 &52.8 &82.6 &89.8 &39.1 &68.6 &77.9 &74.1 &93.0 &96.6 &53.9 &79.4 &87.2 &42.5 &67.5 &78.6 &27.6 &52.5 &63.6 \\
% \hline
SGRAF &79.6 &96.2 &98.5 &63.2 &90.7 &96.1 &57.6 &83.5 &91.5 &41.9 &71.3 &81.2 &77.8 &94.1 &97.4 &58.5 &83.0 &88.8 &43.9 &73.4 &80.1 &28.8 &54.4 &64.2\\
% \hline
GSMN  &78.4 &96.4 &98.6 &63.3 &90.1 &95.7 &55.2 &81.3 &86.2 &37.2 &68.3 &77.3 &76.4 &94.3 &97.3 &57.4 &82.3 &89.0 &43.3 &72.4 &79.3 &26.9 &53.2 &63.8\\
% \hline
VSRN &69.5 &92.3 &97.0 &54.1 &83.4 &90.2 &53.0 &81.1 &89.4 &40.5 &70.6 &81.1 &58.0 &86.1 &91.6 &46.9 &77.0 &85.1 &34.8 &63.1 &73.5 &26.3 &52.5 &64.1 \\
% \hline
NAAF & 78.1 & 96.1 & 98.6 & 63.5 & 89.6 & 95.3 & {58.9} & {85.2} & 92.0 & 42.5 & 70.9 & 81.4 & \underline{79.6} & \underline{96.3} & {98.3} & 59.3 & 83.9 & 90.2 & 44.1 & 69.9 & 78.0 & 31.0 & 54.8 & 64.2 \\
\hline
ALBEF* & 80.1 & 96.9 & 99.0 & 68.3 & 92.5 & 97.1 & \underline{59.7} & 85.8 & 92.3 & \underline{46.1} & 75.8 & 84.9 &63.2 &87.4 &93.5 &48.5 &73.1 & 80.7 &47.7 & 70.4 &79.6 &34.3 &56.3 &65.7\\
% \hline
BLIP*  & 81.4 & 96.7 & {99.2} & 67.7 & 92.1 & 96.3 & 57.8 & 84.1 & 91.2 & 43.4 & 72.8 & 82.7 & 65.0 & 89.5 & 94.1 & 52.2 & 81.9 & 89.2 & 46.2 & 73.2 & 82.2 & 37.2 & 64.0 & 74.5 \\
\hline
CYCLIP* &78.4 &96.0 &98.9 &65.7 &91.5 &96.6 &53.0 &81.0 &89.3 &41.0 &71.7 &82.7 &77.1 &94.1 &97.8 &61.4 &88.1 &92.2 &52.4 &79.7 &87.2 &39.2 &70.5 &79.6\\
% \hline
UMT & \underline{81.5} & {97.3} & 99.1 & {69.2} & {93.2} & {97.2} & 58.4 & 85.0 &{92.7} &{43.6} &{74.5} &{84.3} &77.6 &94.8 &97.9 &{61.9} &{89.8} &{93.3} &52.7 &{81.2} &87.6 &{41.8} &{71.7} &80.4 \\
\hline
X-VLM* &{81.1} & \underline{97.7} & \underline{99.5} & \underline{70.3} & \underline{94.7} & \underline{97.5} & {55.2} & \underline{86.6} & \underline{93.3} & {44.8} & \underline{76.9} & \underline{86.0} & {78.1} & {96.0} & \underline{98.8} & \underline{65.4} & \underline{90.4} & \underline{94.9} & \underline{57.7} & \underline{82.8} & \underline{89.3} & \underline{44.2} & \underline{73.5} & \underline{81.7} \\
X-VLM*+ &\textbf{87.3} &\textbf{98.7} &\textbf{99.8} &\textbf{73.4} &\textbf{94.8} &\textbf{97.7} &\textbf{66.6} &\textbf{90.5} &\textbf{95.5} &\textbf{49.9} &\textbf{79.5} &\textbf{87.6}  &\textbf{85.6} &\textbf{98.1} &\textbf{99.3} &\textbf{73.3} &\textbf{93.0} &\textbf{96.1} &\textbf{63.9} &\textbf{85.7} &\textbf{90.6} &\textbf{50.7} &\textbf{75.5} &\textbf{83.7}\\
\hline
\end{tabular*}}
\end{table*}

\begin{table*}[htbp]{
\centering
\caption{Single-modal retrieval performance comparison. Evaluation criteria is NDCG@A. The method with “+" sign is our method.}
\label{tab:tab2}
\begin{tabular*}{0.985\textwidth}{@{\extracolsep{\fill}}@{}l@{\hspace{0.06cm}}|@{}c@{\hspace{0.06cm}}|@{}c@{\hspace{0.06cm}}|@{}c@{\hspace{0.06cm}}|@{}c@{\hspace{0.06cm}}|@{}c@{\hspace{0.06cm}}|@{}c@{\hspace{0.06cm}}|@{}c@{\hspace{0.06cm}}|@{}c@{\hspace{0.06cm}}|@{}c@{\hspace{0.06cm}}|@{}c@{\hspace{0.06cm}}|@{}c@{\hspace{0.06cm}}|@{}c@{\hspace{0.06cm}}|@{}c@{\hspace{0.06cm}}|@{}c@{\hspace{0.06cm}}|@{}c@{\hspace{0.06cm}}|@{}c@{\hspace{0.06cm}}|@{}c@{\hspace{0.06cm}}|@{}c@{\hspace{0.06cm}}|@{}c@{\hspace{0.06cm}}|@{}c@{\hspace{0.06cm}}|@{}c@{\hspace{0.06cm}}|@{}c@{\hspace{0.06cm}}|@{}c@{\hspace{0.06cm}}|@{}c@{\hspace{0.06cm}}}
\hline
\multirow{3}{*}{Methods} & \multicolumn{6}{c|}{MS-COCO (1K)} & \multicolumn{6}{c|}{MS-COCO (5K)} & \multicolumn{6}{c|}{FLICKR30K} & \multicolumn{6}{c}{Vizwiz}\\
\cline{2-25}
& \multicolumn{3}{c|}{I2I} & \multicolumn{3}{c|}{T2T} 	& \multicolumn{3}{c|}{I2I} & \multicolumn{3}{c|}{T2T} & \multicolumn{3}{c|}{I2I} & \multicolumn{3}{c|}{T2T} & \multicolumn{3}{c|}{I2I} & \multicolumn{3}{c}{T2T}\\
\cline{2-25}
& $@$10 & $@$20 & $@$50  & $@$10 & $@$20 & $@$50 & $@$10 & $@$20 & $@$50 & $@$10 & $@$20 & $@$50 & $@$10 & $@$20 & $@$50 & $@$10 & $@$20 & $@$50 & $@$10 & $@$20 & $@$50 & $@$10 & $@$20 & $@$50\\
\hline
VSE++ &62.5 & 64.2 & 66.5 & 67.9 & 69.4 & 70.2 & 43.2 & 45.2 & 48.5 & 43.8 & 45.5 & 48.6 & 59.0 & 60.9 & 64.0 & 63.3 & 64.0 & 64.2 & 54.2 & 55.6 & 56.7 & 57.8 & 58.6 & 59.3\\
% \hline
SCAN  &64.1 & 65.1 & 66.7 & 69.9 & 70.3 & 69.5 & 59.7 & 61.2 & 63.4 & 63.5 & 64.8 & 65.8 &52.6 &55.0 &59.1 & 38.7 &42.6 &47.6 & 55.4 &57.4 &61.1 &52.0 &54.0 & 56.5 \\
% \hline
IMRAM &65.1 & 66.6 & 68.6 & 70.2 & 70.5 & 70.1 & 60.4 & 62.0 & 64.5 & {64.4} & {66.1} & {67.9} & 61.1 & 62.8 & 65.7 & 66.8 & 66.4 & 65.5 & 59.5 & 61.5 & 65.0 & 61.2 & 62.4 & {63.1}\\
% \hline
SGRAF &62.5 & 64.0 & 66.3 & 61.3 & 63.4 & 65.0 & 56.7 & 58.3 & 60.8 & 53.8 & 56.1 & 58.4 & 58.7 & 60.8 & 64.3 & 57.7 & 60.2 & 62.9 & 59.1 & 61.1 & 64.7 & 56.3 & 59.3 & 62.9\\
% \hline
GSMN  &61.6 & 63.3 & 65.9 & 59.7 & 61.1 & 61.5 & 55.1 & 56.7 & 59.1 & 49.1 & 50.7 & 53.1 & 60.1 & 61.9 & 65.2 & 53.6 & 56.4 & 59.7 & 56.4 & 57.3 & 58.6 & 52.5 & 56.3 & 55.9\\
% \hline
VSRN &52.9 &55.1 &58.8 &58.3 &57.6 &56.9 & 45.0 & 46.3 & 49.1 &60.3 & 58.4 & 56.3 & 61.7 & 63.4 & 66.4 & {67.8} & {67.4} & {66.3} & 62.0 &63.7 &66.7 & 59.8 &60.8 &61.6\\
% \hline
NAAF  & 63.7 & 64.8 & 66.8 & {70.7} & {70.9} & {72.1} & 59.5 & 60.8 & 62.8 & {65.3} & {66.6} & \textbf{68.6} & 60.5 & 62.2 & 65.2 & {71.1} & {69.9} & {68.2} & 59.1 & 61.2 & 64.7 & 61.7 & 61.8 & 62.9 \\
\hline
ALBEF* &68.2 & 69.1 & 70.9 & 49.6 & 50.6 & 51.5 & 64.3 & 66.1 & 68.3 & 47.5 & 49.7 & 51.1 &60.8 &62.3 &65.2 &70.0 &68.4 & 66.2 &61.8 & 63.3 &66.2 &63.7 &63.0 &62.1\\
% \hline
BLIP* & {68.1} & {68.9} & 70.1 & 64.4 & 63.5 & 62.4 & 64.0 & 65.5 & 67.6 & 59.0 & 59.3 & 59.5 & 62.6 & 64.2 & 67.0 & 65.7 & 64.4 & 62.9 & 63.0 & 64.9 & \underline{68.0} & 58.3 & 58.8 & 59.6 \\
\hline
CYCLIP* &67.3 &68.2 &69.0 &68.9 &68.4 &67.3 &62.7 &64.0 &66.3 &63.4 &64.5 &66.8 &61.9 &63.3 &66.2 &63.4 &64.0 &63.9 &63.0 &64.5 &67.4 &56.0 &58.2 &59.9\\
% \hline
UMT &67.5 &68.8 &{70.7} &68.1 &65.9 &64.4 &{65.2} &\textbf{68.0} &67.7 &57.3 &57.2 &66.3 &62.7 &64.2 &66.9 &66.8 &65.3 &63.9 & \underline{64.0} &\underline{65.5} &67.4 &59.0 &60.2 &61.9 \\
\hline
X-VLM* &\underline{69.2} &\underline{70.0} &\underline{71.1} &\underline{72.8} &\underline{72.9} &\underline{72.3} &\underline{65.3} &{66.8} &\underline{68.8} &\underline{65.5} &\underline{66.7} &67.8 & \underline{64.1} & \underline{65.2} & \underline{67.7} & \underline{72.3} & \underline{70.9} & \underline{68.7} & \textbf{64.4} & \textbf{66.1} &\textbf{68.9} & \underline{64.8} & \underline{64.9} & \underline{65.0}\\
X-VLM*+ &\textbf{70.1} &\textbf{70.9} &\textbf{72.1} &\textbf{73.5} &\textbf{73.4} &\textbf{72.5} &\textbf{66.3} &\underline{67.8} &\textbf{69.8} &\textbf{65.8} &\textbf{67.0} &\underline{68.0}  &\textbf{64.3} &\textbf{65.4} &\textbf{67.8} &\textbf{73.0} &\textbf{71.4} &\textbf{68.9} &\textbf{64.4} &\textbf{66.1} &\textbf{68.9} &\textbf{67.2} &\textbf{66.5} &\textbf{65.5}\\
\hline
\textbf{SWIN} &66.5 &67.7 &69.1 &- &- &- &61.5 &63.2 &65.7 &- &- &- &64.1 &64.5 &67.9 &- &- &- &50.3 &53.9 &57.3 &- &- &-\\	
% \hline
\textbf{BERT} &- &- &- &71.1 &71.1 &70.2 &- &- &- &63.4 &64.8 &66.1 &- &- &- &71.6 &70.2 &68.4 &- &- &- &55.5 &57.5 &59.6 \\
\hline
\end{tabular*}}
\end{table*}

\begin{figure*}[t]
    \centering
    \begin{minipage}[b]{0.24\textwidth}
        \centering
        \includegraphics[width=\textwidth]{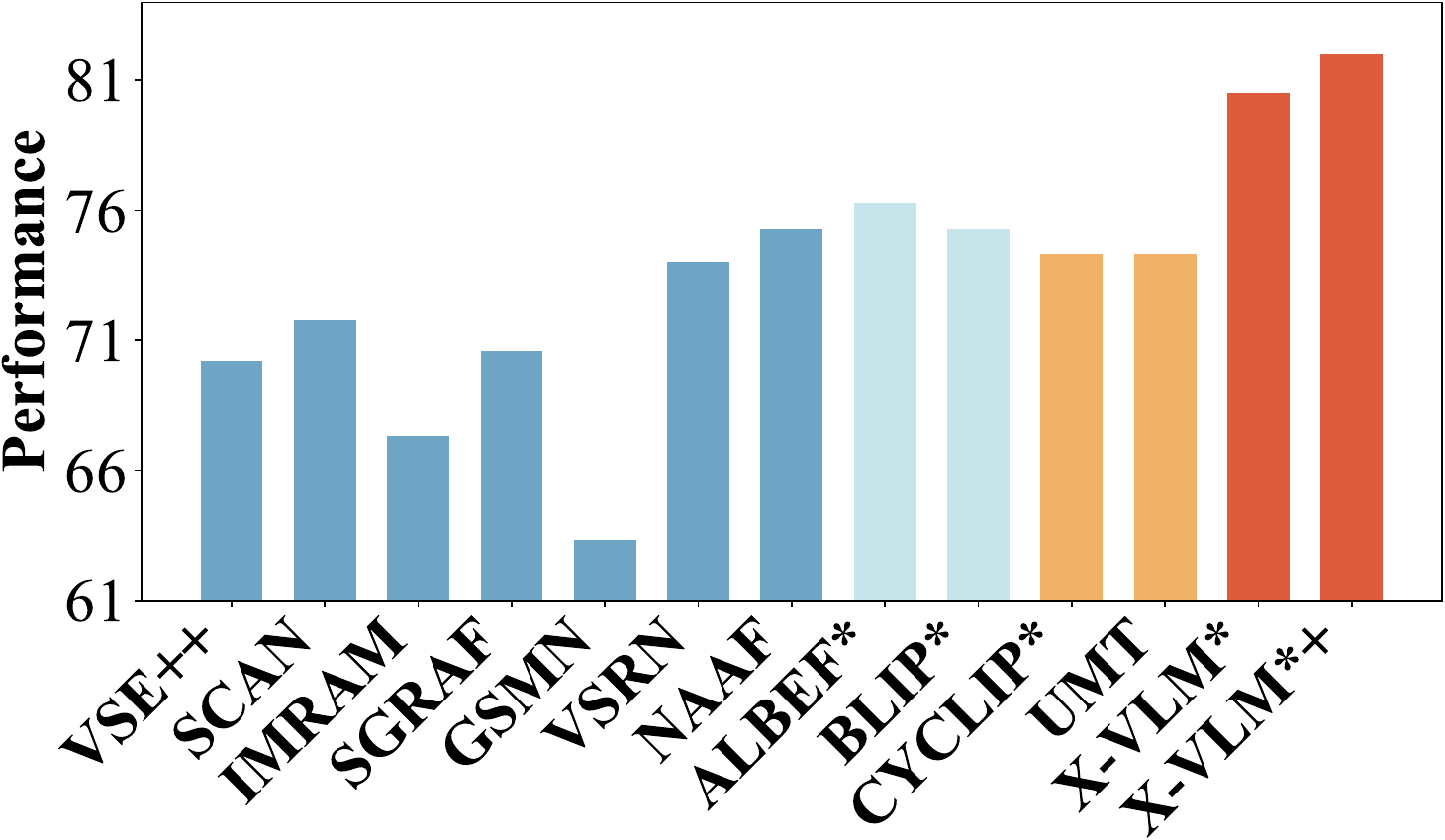}
        \small (a) MS-COCO (1K).
    \end{minipage}
    \hfill
    \begin{minipage}[b]{0.24\textwidth}
        \centering
        \includegraphics[width=\textwidth]{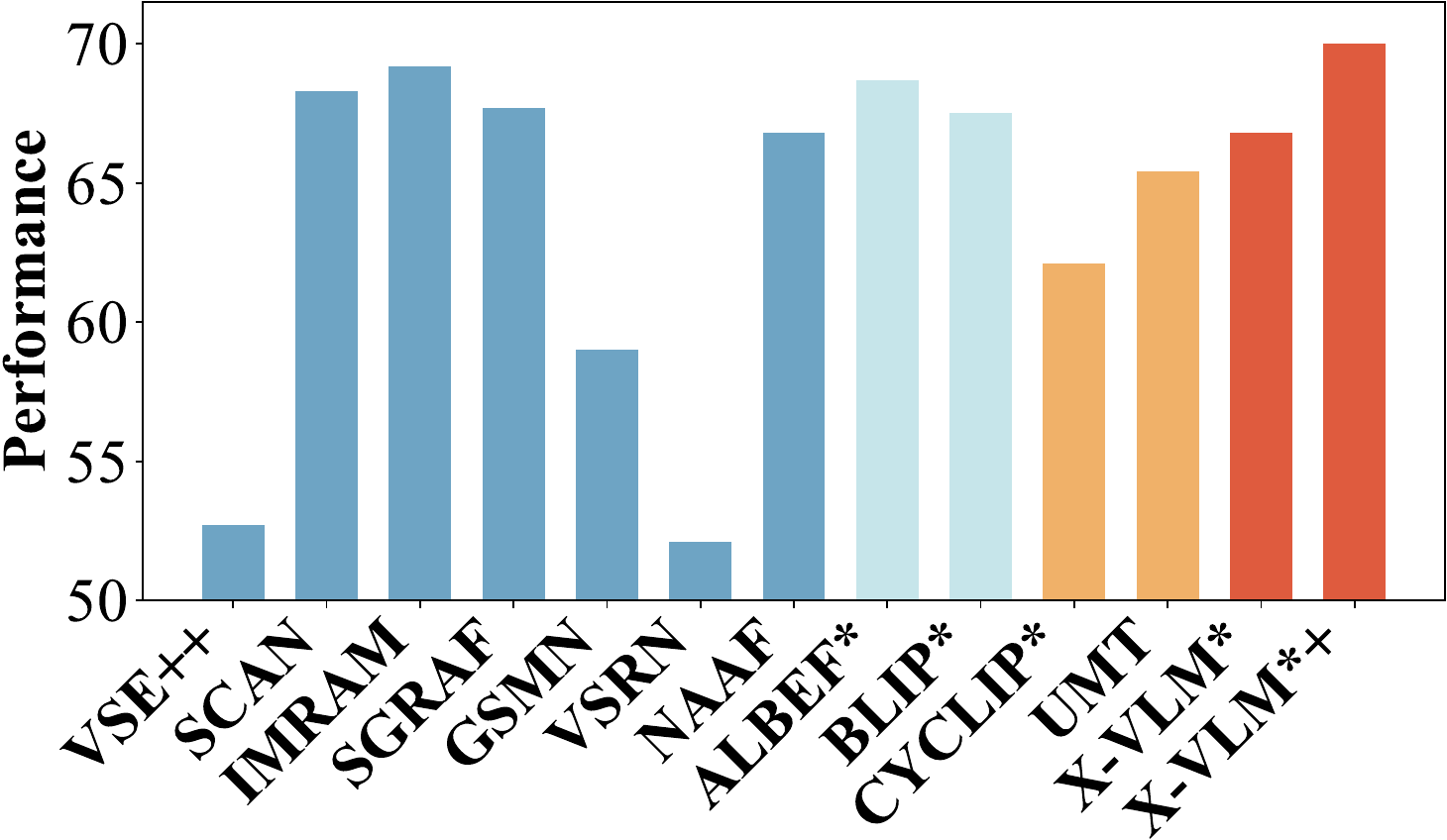}
        \small (b) MS-COCO (5K).
    \end{minipage}
    \hfill
    \begin{minipage}[b]{0.24\textwidth}
        \centering
        \includegraphics[width=\textwidth]{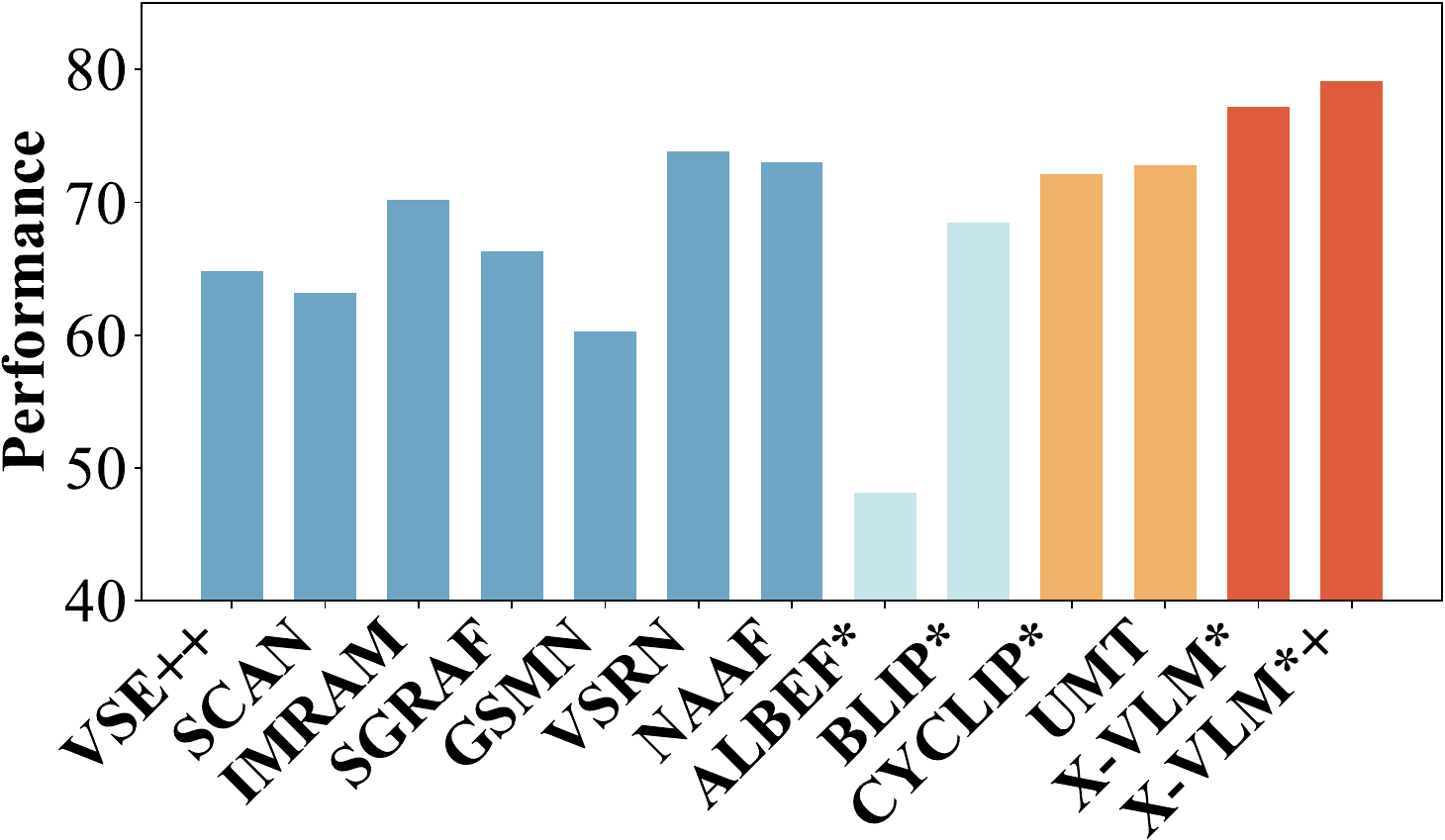}
        \small (c) FLICKR30K.
    \end{minipage}
    \hfill
    \begin{minipage}[b]{0.24\textwidth}
        \centering
        \includegraphics[width=\textwidth]{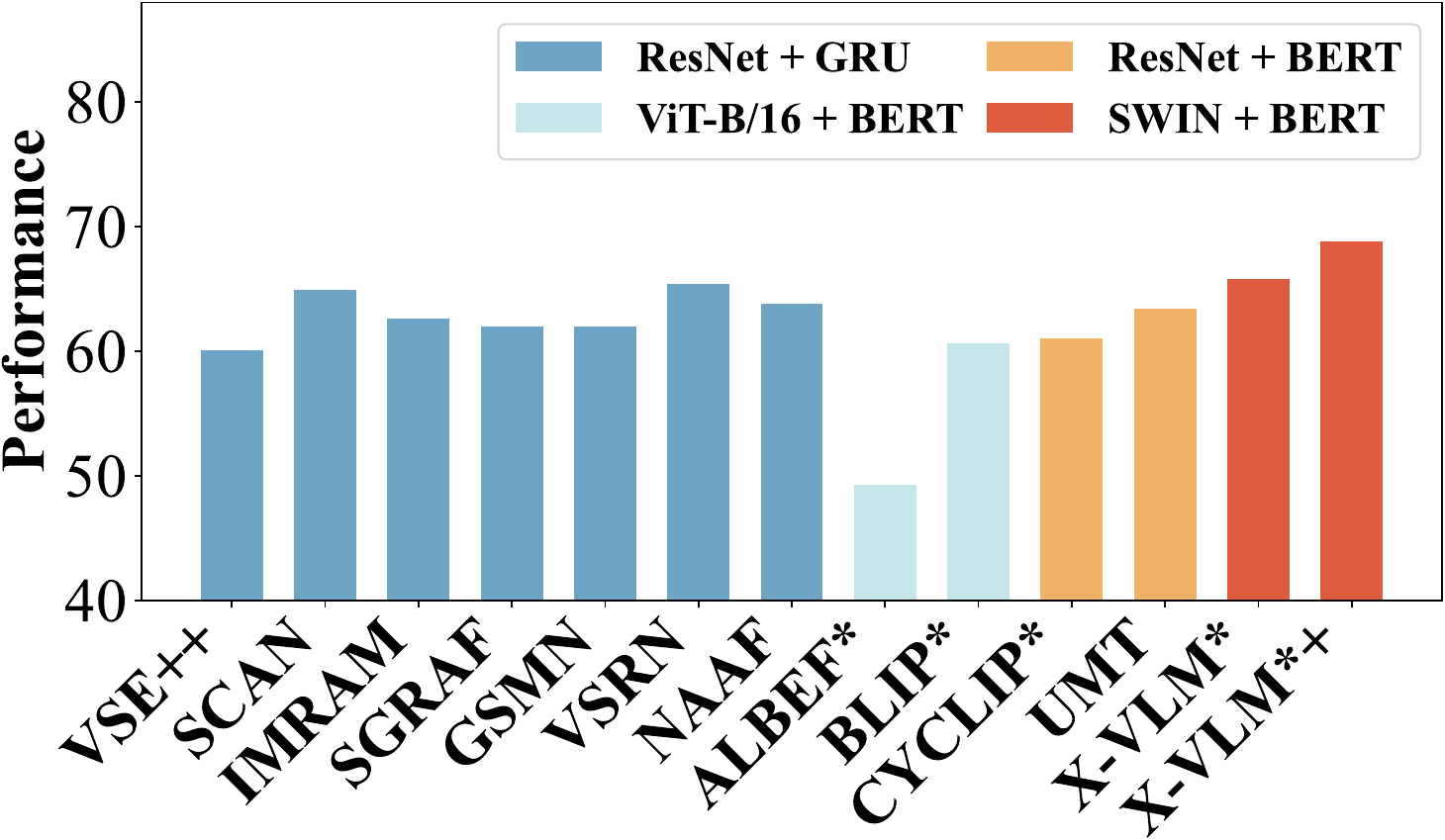}
        \small (d) Vizwiz.
    \end{minipage}
    % \caption{The results of I2IT$@10$ retrieval.}\label{fig:I2IT@10}
    \begin{minipage}[b]{0.24\textwidth}
        \centering
        \includegraphics[width=\textwidth]{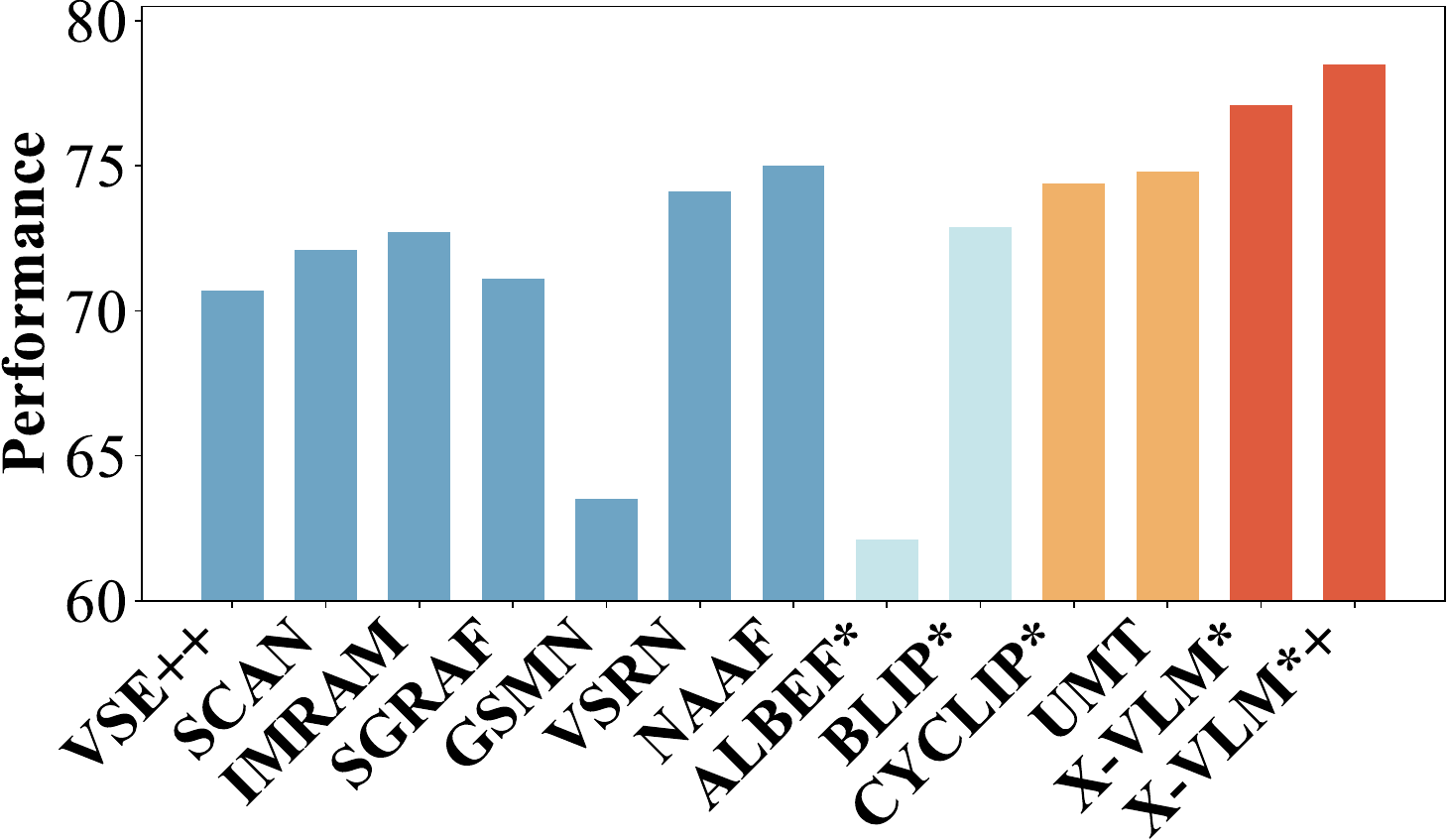}
        \small (e) MS-COCO (1K).
    \end{minipage}
    \hfill
    \begin{minipage}[b]{0.24\textwidth}
        \centering
        \includegraphics[width=\textwidth]{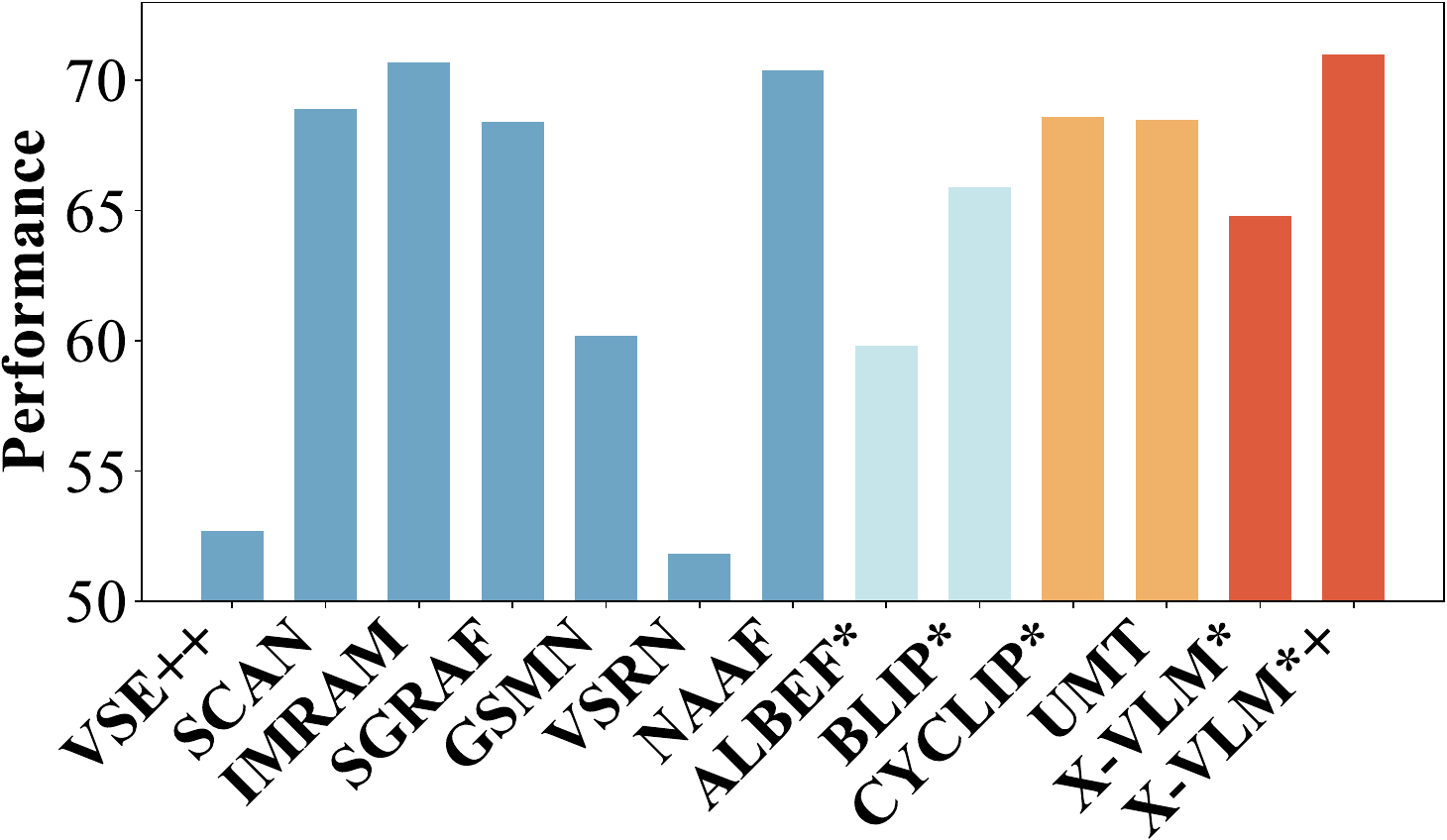}
        \small (f) MS-COCO (5K).
    \end{minipage}
    \hfill
    \begin{minipage}[b]{0.24\textwidth}
        \centering
        \includegraphics[width=\textwidth]{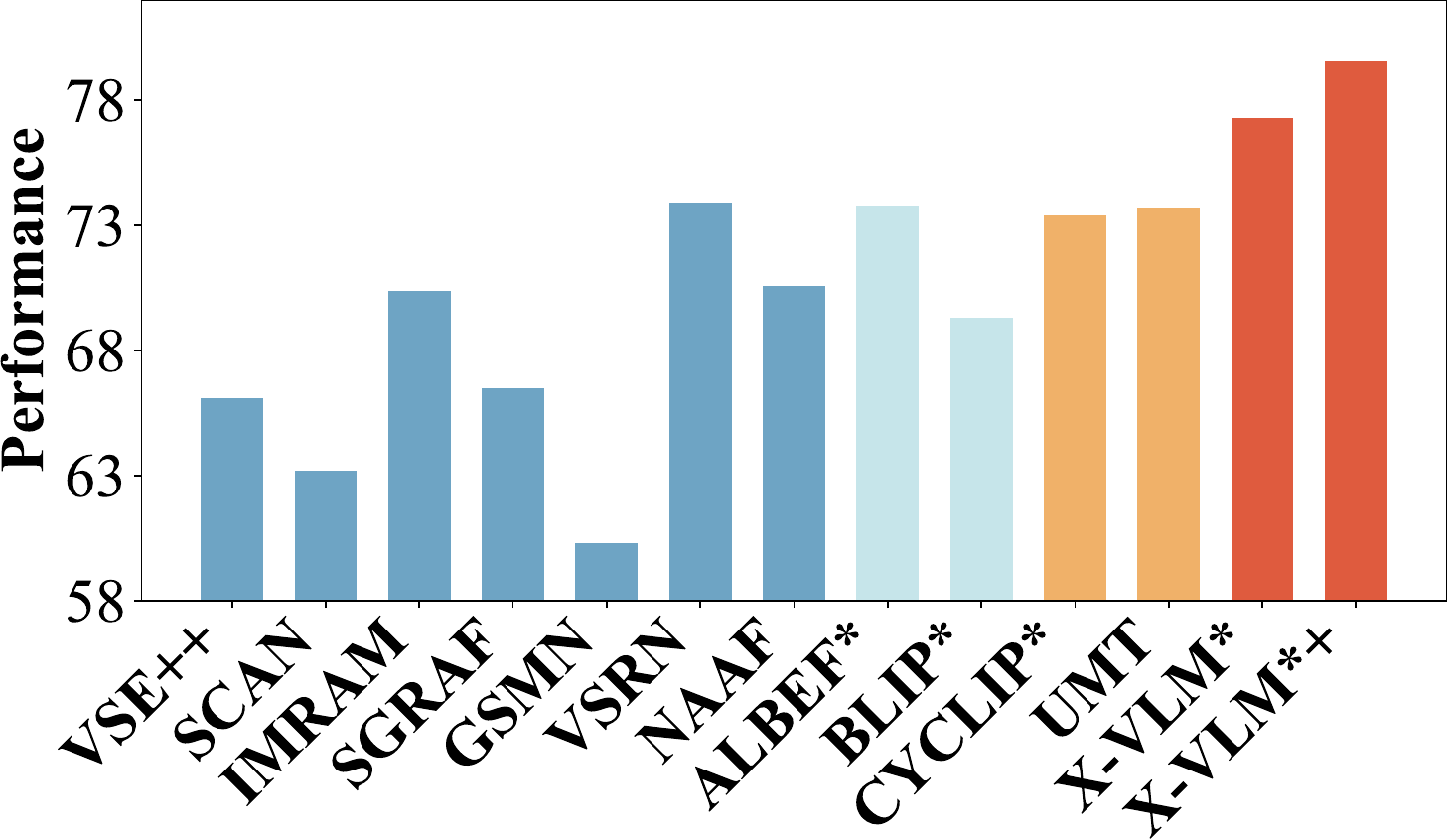}
        \small (g) FLICKR30K.
    \end{minipage}
    \hfill
    \begin{minipage}[b]{0.24\textwidth}
        \centering
        \includegraphics[width=\textwidth]{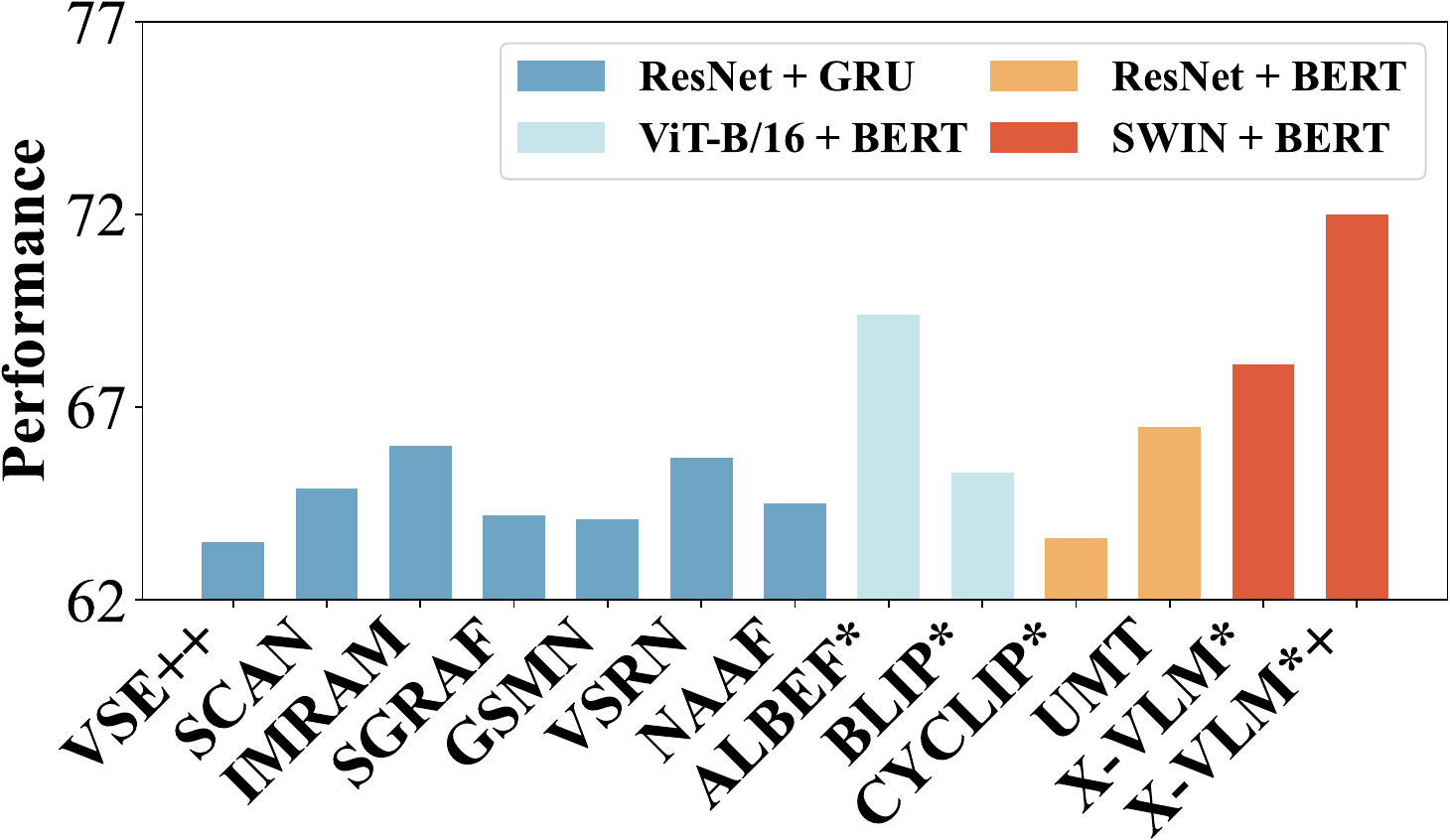}
        \small (h) Vizwiz.
    \end{minipage}
    % \caption{The results of T2IT$@10$ retrieval.}\label{fig:T2IT@10}
    \caption{The results of I2IT$@10$ (a--d), T2IT$@10$ (e--h) of mixed retrieval task. The method with “+" sign, i.e., X-VLM*+, is our method.}\label{fig:single_mixed}
\end{figure*}

\noindent {\bf{Structure-Aware Distillation:}} As shown in Fig.~\ref{fig:fra}, we calculate the relational similarity matrices $S_I, S_T, S_{IT}$ for two modal independent networks and the cross-modal network, by computing the distances of data examples in the mini-batch. Using the $S_I$ as an example: $S_I(i, j) = \cos(I_{i_{CLS}}, I_{j_{CLS}})$. Consequently, we can acquire the optimal teacher matrix by fusing two teachers' matrices $S_{O} = \lambda S_I + (1-\lambda) S_T$, where $\lambda$ is a learnable parameter aimed at finding the optimal modality fusion. For measuring the teacher-student relational consistency, we adopt the Mean Absolute Error (MAE) between two matrices, which aims to find subtler differences by considering matrices as high-dimensional objectives rather than simple aggregations. Consequently, given $S_O, S_{IT}$, the structure-aware distillation is defined as:
\begin{equation}\nonumber
\begin{matrix}
\ell_{sa} = \frac{1}{J} \sum_{m=1}^J \sum_{n=1, n \neq m}^{J}|S_O(m, n)-S_{IT}(m, n)|.
\end{matrix}
\end{equation} 
This effectively alleviates the disruption of structural information during modal alignment. Therefore, the overall multi-granularity distillation can be represented as: $\ell_{md} = \ell_{iic} + \ell_{ttc} + \ell_{sa}$.

In summary, we combine cross-modal matching loss and distillation objectives to study the common cross-modal representations, which can reduce the cross-modal gap and preserve structure simultaneously. The overall loss can be formulated as: $L = \ell_{cr} + \ell_{md}$.

\section{Experiments} 
\subsection{Data Description}
To verify the effectiveness of our method, we conduct experiments on three datasets. In detail, \textbf{MS-COCO}~\cite{LinMBHPRDZ14} contains 123,287 images, including 82,783 training images and 40,504 validation images, each labeled with 5 captions. Following~\cite{KarpathyL15}, we use the splits of 5,000 images for validation, 1,000/5,000 images for testing, and the rest for training. \textbf{FLICKR30K}~\cite{HuiskesL08} consists of 31,000 images, and each image is associated with 5 captions. The dataset is split into 29,000 training images, 1,000 validation images, and 1,000 testing images following~\cite{KarpathyL15}. \textbf{Vizwiz}~\cite{GurariZZB20} consists of 39,181 images originating from people who are blind that are each paired with 5 captions. The dataset is split into 23,431 training instances, 7,750 validation instances, and 8,000 testing instances.

\subsection{Baselines and Evaluation Protocol}
For comparison methods, we evaluate four types of state-of-the-art approaches: 1). Dual-encoder  retrieval methods, including VSE++~\cite{FaghriFKF18}, SCAN~\cite{LeeCHHH18}, IMRAM~\cite{ChenDLLLH20}, SGRAF~\cite{DiaoZML21}, GSMN~\cite{LiuMZX0Z20}, VSRN~\cite{li2019visual}, and NAAF~\cite{NAAF}. 2). Transformer-based retrieval methods, including ALBEF~\cite{Li2021}, BLIP~\cite{DBLP:conf/icml/0001LXH22}, and X-VLM~\cite{Hang2021}. 3). Single-modal models, e.g., Swin Transformer~\cite{LiuL00W0LG21} and BERT~\cite{DevlinCLT19}. 4). Imbalance multi-modal learning methods, i.e., CYCLIP~\cite{abs-2205-14459} and UMT~\cite{abs-2106-11059}. Note that CYCLIP incorporates regularizers about single-modal and cross-modal structural information into the cross-modal contrastive item in the form of a multi-task loss, and UMT optimizes the cross-modal model by distilling the instance' relations learned by the single-modal model. Since the baselines utilize different backbones, we conducted an experiment that integrates our method with these baselines in a plug-and-play manner to ensure a fair comparison. The detailed results are provided in Section~\uppercase\expandafter{\romannumeral3} of the supplementary.

In our experiments, we focus on three tasks: 1). cross-modal retrieval, 2). single-modal retrieval, and 3). mixed retrieval. Detailed implementation of these three retrieval tasks is provided in Section~\uppercase\expandafter{\romannumeral1} of the supplementary. We evaluate performance using the recall at A (R$@$A) metric, which is commonly adopted in most cross-modal retrieval methods for both image-to-text (I2T) and text-to-image (T2I) retrieval tasks~\cite{FaghriFKF18, MessinaAEFGM21, Li2021, Hang2021}. During cross-modal retrieval, the corresponding captions of the given image or corresponding images of the given caption are expected, while in single-modal retrieval, what we need are relatively similar instances~\cite{tang2015neighborhood, wang2022learning, DBLP:journals/tip/GaoTHYDZC12}. As a result, the R$@$A metric as a binary correlation that only examines whether the retrieval results are relevant is not suitable for single-modal and mixed retrieval (I2IT and T2IT). Following~\cite{MessinaAEFGM21, CarraraEFFF18}, a more comprehensive metric NDCG$@$A ~\cite{CarraraEFFF18} is adopted instead to promote the items with higher relevance scores to appear in better ranking positions. Note that the ranking is computed using text similarity scores (i.e., ROUGE-L~\cite{HuangWCW19}) between a sentence and the sentences associated with a certain image.

\subsection{Implementation Details}
Our multi-granularity distillation approach can be seen as a plug-and-play module, and we can incorporate other cross-modal retrieval models as student models to validate the generalization capability of the multi-granularity distillation. Without any loss of generality, we choose X-VLM without the bounding box module as the student model to validate the effectiveness of our method, and in subsequent experiments, we will extend this module to traditional cross-modal retrieval and visual-language pre-training models, to further demonstrate its generalization. X-VLM consists of a vision encoder which is initialized using Swin Transformer/12~\cite{DosovitskiyB0WZ21}, a language encoder which is initialized using the first 6 layers of the BERTbase~\cite{DevlinCLT19}, and text-oriented cross-attention transformer are initialized using the last 6 layers of the BERTbase. 
In total, our model has 214.4M parameters for training. We also study the impact of using larger models such as CLIP for training, as discussed in Section~\uppercase\expandafter{\romannumeral5} of the supplementary. The images' resolution is 384$\times$384 as input. For text input, the maximum number of tokens is 30. We use the AdamW~\cite{loshchilov2017decoupled} optimizer with a weight decay of 0.1. The learning rate is warmed up to $0$ from $3e-5$ in the first epoch and decayed to $1e-5$ following a cosine schedule. The momentum parameter for updating the momentum model is 0.995. The model is trained for 10 epochs with a total batch size of 36 on 6 NVIDIA A6000 GPUs.

\begin{table}[htbp]{
    \centering
    \caption{Ablation studies performance comparison. \\Evaluation criteria are R$@$A and NDCG$@$A.}
    \label{tab:tab4}
    \begin{tabular*}{0.49\textwidth}{@{\extracolsep{\fill}}@{}l@{}|@{}c@{}|@{}c@{}|@{}c@{}|@{}c@{}|@{}c@{}|@{}c@{}|@{}c@{}|@{}c@{}|@{}c@{}|@{}c@{}|@{}c@{}|@{}c@{}}
    % \begin{tabular*}{0.985\textwidth}{@{\extracolsep{\fill}}@{}l@{\hspace{0.06cm}}|@{}c@{\hspace{0.06cm}}|@{}c@{\hspace{0.06cm}}|@{}c@{\hspace{0.06cm}}|@{}c@{\hspace{0.06cm}}|@{}c@{\hspace{0.06cm}}|@{}c@{\hspace{0.06cm}}|@{}c@{\hspace{0.06cm}}|@{}c@{\hspace{0.06cm}}|@{}c@{\hspace{0.06cm}}|@{}c@{\hspace{0.06cm}}|@{}c@{\hspace{0.06cm}}|@{}c@{\hspace{0.06cm}}}
    \hline
    \multirow{3}{*}{Methods} &\multicolumn{6}{c|}{FLICKR30K} & \multicolumn{6}{c}{Vizwiz}\\
    \cline{2-13}
    & \multicolumn{3}{c|}{I2T} & \multicolumn{3}{c|}{T2I} & \multicolumn{3}{c|}{I2T} & \multicolumn{3}{c}{T2I}\\
    \cline{2-13}
    & $@$1 & $@$5 & $@$10 & $@$1 & $@$5 & $@$10 & $@$1 & $@$5 & $@$10 & $@$1 & $@$5 & $@$10\\
    \hline
    {w/o $\ell_{itc}$} &\underline{84.8} &\underline{97.6} &\textbf{99.3} &\underline{71.4} &\underline{92.6} &\underline{95.7} &\underline{61.4} &\underline{85.4} &90.3 &\underline{47.7} &75.1 &83.1 \\
    {w/o $\ell_{itm}$} &78.6 &97.0 &98.9 &65.7 &90.3 &94.5 &60.2 &85.3 &\textbf{90.8} &46.3 &74.7 &83.1 \\
    {w/o $\ell_{iic}$} &83.5 &96.9 &98.8 &68.0 &91.6 &95.3 &59.4 &83.2 &89.8 &45.1 &73.6 &81.9 \\
    {w/o $\ell_{ttc}$} &84.1 &97.1 &\underline{99.1} &67.7 &91.6 &94.8 &61.1 &85.0 &90.4 &46.9 &\underline{75.2} &\underline{83.4} \\
    {w/o $\ell_{sa}$} &81.2 &95.8 &99.0 &67.3 &90.9 &95.0 &58.1 &82.8 &89.9 &45.2 &73.9 &82.3 \\
    {Ours} &\textbf{85.6} &\textbf{98.1} &\textbf{99.3} &\textbf{73.3} &\textbf{93.0} &\textbf{96.1} &\textbf{63.9} &\textbf{85.7} &\underline{90.6} &\textbf{50.7} &\textbf{75.5} &\textbf{83.7}\\
    \hline
    & \multicolumn{3}{c|}{I2I} & \multicolumn{3}{c|}{T2T} & \multicolumn{3}{c|}{I2I} & \multicolumn{3}{c}{T2T}\\
    \cline{2-13}
    & $@$10 & $@$20 & $@$50 & $@$10 & $@$20 & $@$50 & $@$10 & $@$20 & $@$50 & $@$10 & $@$20 & $@$50\\
    \hline
    {w/o $\ell_{itc}$} &63.1 &64.5 &67.2 &72.0 &70.9 &68.4 &63.1 &64.8 &67.5 &65.3 &64.9 &64.6 \\
    {w/o $\ell_{itm}$} &\textbf{64.3} &\textbf{65.6} &\textbf{68.2} &\textbf{73.7} &\textbf{72.1} &\textbf{69.9} &64.0 &65.7 &68.5 &\underline{66.0} &\underline{65.8} &\underline{65.4} \\
    {w/o $\ell_{iic}$} &63.9 &65.0 &67.2 &72.9 &70.9 &68.1 &63.8 &65.8 &68.4 &64.4 &64.8 &65.2 \\
    {w/o $\ell_{ttc}$} &\underline{64.1} &65.0 &67.0 &72.7 &70.3 &67.2 &\underline{64.2} &\underline{65.9} &\underline{68.7} &65.3 &65.4 &65.3 \\
    {w/o $\ell_{sa}$} &63.5 &64.7 &66.9 &72.7 &70.0 &67.3 &64.0 &65.5 &68.1 &64.6 &65.1 &\underline{65.4} \\
    {Ours} &\textbf{64.3} &\underline{65.4} &\underline{67.8} &\underline{73.0} &\underline{71.4} &\underline{68.9} &\textbf{64.4} &\textbf{66.1} &\textbf{68.9} &\textbf{67.2} &\textbf{66.5} &\textbf{65.5}\\
    \hline
    & \multicolumn{3}{c|}{I2IT} & \multicolumn{3}{c|}{T2IT} & \multicolumn{3}{c|}{I2IT} & \multicolumn{3}{c}{T2IT}\\
    \cline{2-13}
    & $@$10 & $@$20 & $@$50 & $@$10 & $@$20 & $@$50 & $@$10 & $@$20 & $@$50 & $@$10 & $@$20 & $@$50\\
    \hline
    {w/o $\ell_{itc}$} &78.7 &73.9 &68.3 &77.2 &75.3 &72.3 &66.8 &65.2 &62.3 &68.5 &67.8 &66.2 \\
    {w/o $\ell_{itm}$} &\textbf{82.7} &\textbf{79.0} &\textbf{74.4} &\textbf{82.9} &\textbf{79.2} &\textbf{74.5} &\textbf{75.1} &\textbf{74.8} &\textbf{71.0} &\textbf{78.3} &\textbf{75.0} &\textbf{71.2} \\
    {w/o $\ell_{iic}$} &78.0 &73.7 &68.4 &77.3 &75.3 &72.0 &66.8 &65.6 &63.0 &68.4 &68.0 &67.0 \\
    {w/o $\ell_{ttc}$} &78.2 &73.5 &68.2 &76.4 &75.7 &72.2 &67.0 &66.0 &62.5 &67.7 &67.6 &66.5 \\
    {w/o $\ell_{sa}$} &77.7 &73.3 &67.3 &77.1 &74.8 &71.9 &66.3 &65.1 &62.8 &68.5 &68.1 &67.1 \\
    {Ours} &\underline{79.1} &\underline{74.0} &\underline{68.8} &\underline{79.6} &\underline{76.7} &\underline{72.5} &\underline{68.8} &\underline{66.3} &\underline{63.1} &\underline{72.0} &\underline{70.2} &\underline{67.6}\\
    \hline
    \end{tabular*}
}
\end{table}

\subsection{Retrieval Results}
\noindent {\bf{Cross-Modal Retrieval:}} I2T and T2I retrievals are considered to evaluate cross-modal retrieval performance. Table \ref{tab:tab1} records the results on four public datasets. To ensure experimental fairness, we exclusively utilized the given datasets for model training, rather than pre-training additional data as the large-scale models (i.e., ALBEF, X-VLM, CYCLIP, and BLIP) do. So we retrained the ALBEF  and other large models from scratch (marked with ``*''), thus causing different results compared to the original paper. Results indicate: 1). Our X-VLM*+ outperforms the best cross-modal retrieval method, i.e., X-VLM, on I2T Recall$@$1 and T2I Recall$@$1 by 6.2/11.4/7.5/6.2 and 3.1/5.1/7.9/6.5 respectively, on four datasets. This phenomenon reveals that structure preservation can also enhance the learning of cross-modal consistent representations by bringing image (i.e., weak modality) representations closer to the text (i.e., strong modality) ones without compromising the original single-modal structure. 2). Our X-VLM*+ outperforms CYCLIP* (multi-task optimization) in all settings, which emphasizes that multi-granularity distillation emerges as a relatively superior strategy. 3). The performance of large-scale pre-trained models is limited when data is limited, e.g., ALBEF* and X-VLM* exhibit competitive performance with dual-encoder models like SGRAF and NAAF.

% 4.6 Parameter analyses
\begin{figure*}[!htb]{\small
\begin{minipage}[h]{58mm}
\centering
\includegraphics[width=58mm]{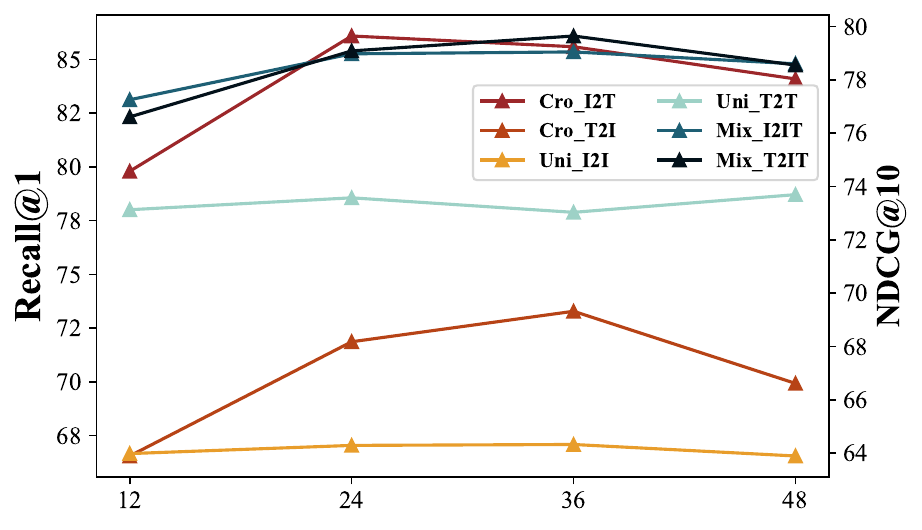}\\
\mbox{  ({\it a}) Batch Size analysis (FLICKR30K).}
\end{minipage}
\begin{minipage}[h]{58mm}
\centering
\includegraphics[width=58mm]{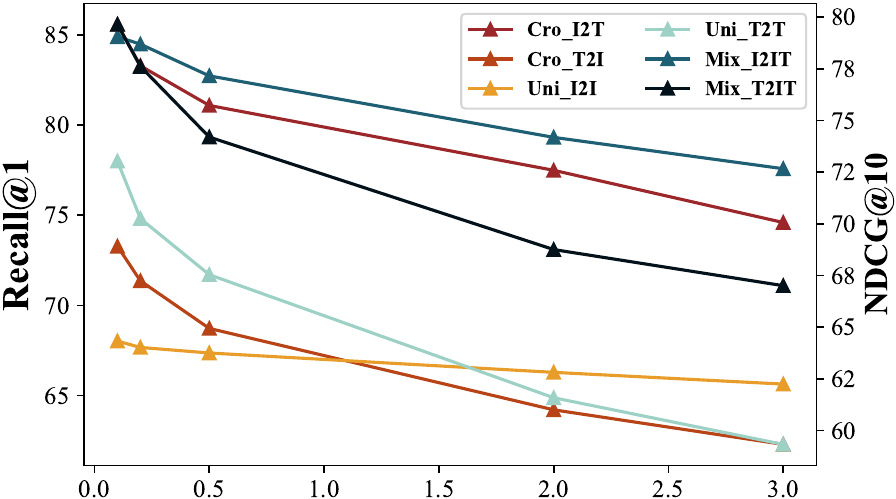}\\
\mbox{  ({\it b}) $\tau$ analysis (FLICKR30K).}
\end{minipage}
\begin{minipage}[h]{58mm}
\centering
\includegraphics[width=58mm]{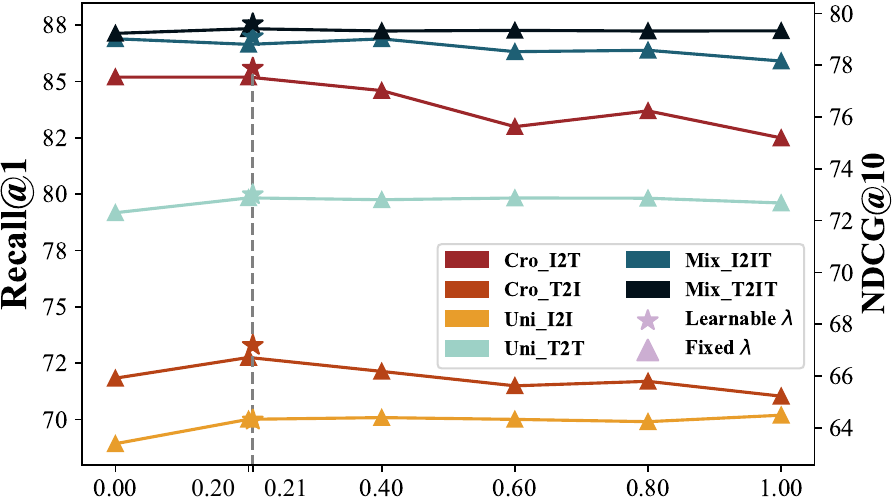}\\
\mbox{  ({\it c}) $\lambda$ analysis (FLICKR30K).}
\end{minipage}\\
\begin{minipage}[h]{58mm}
\centering
\includegraphics[width=58mm]{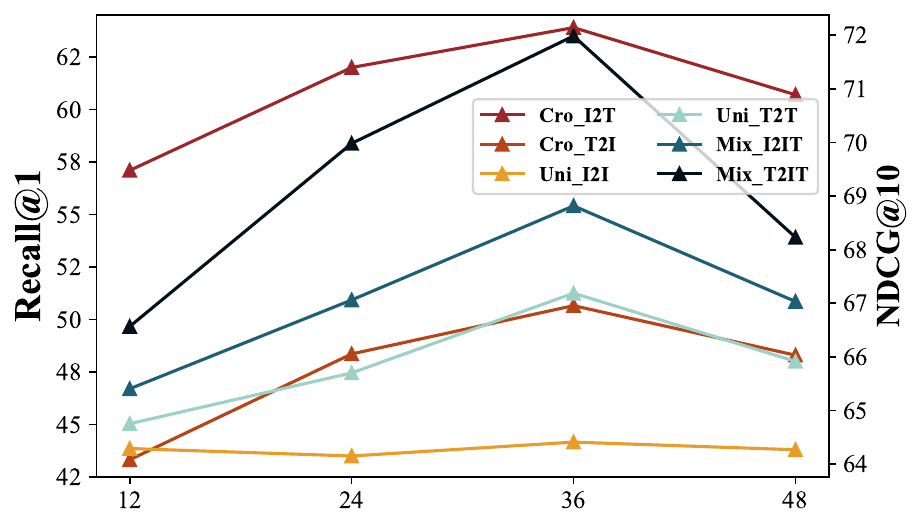}\\
\mbox{  ({\it d}) Batch Size analysis (Vizwiz).}
\end{minipage}
\begin{minipage}[h]{58mm}
\centering
\includegraphics[width=58mm]{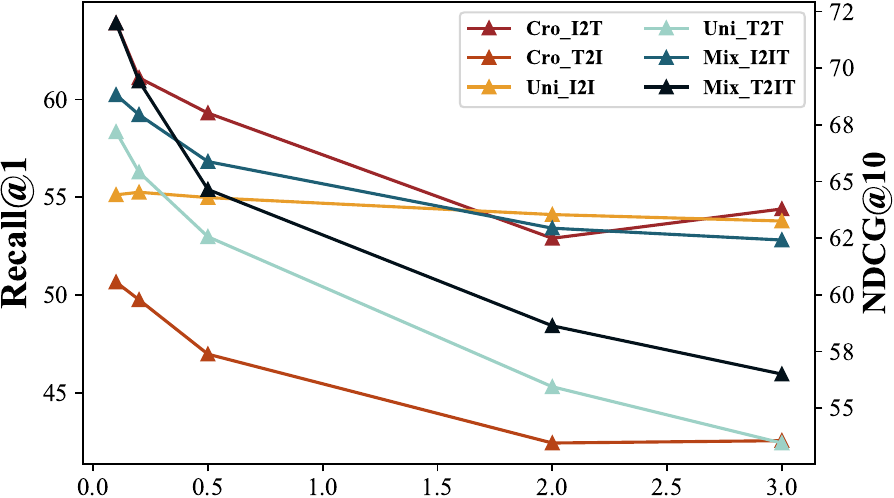}\\
\mbox{  ({\it e}) $\tau$ analysis (Vizwiz).}
\end{minipage}
\begin{minipage}[h]{58mm}
\centering
\includegraphics[width=58mm]{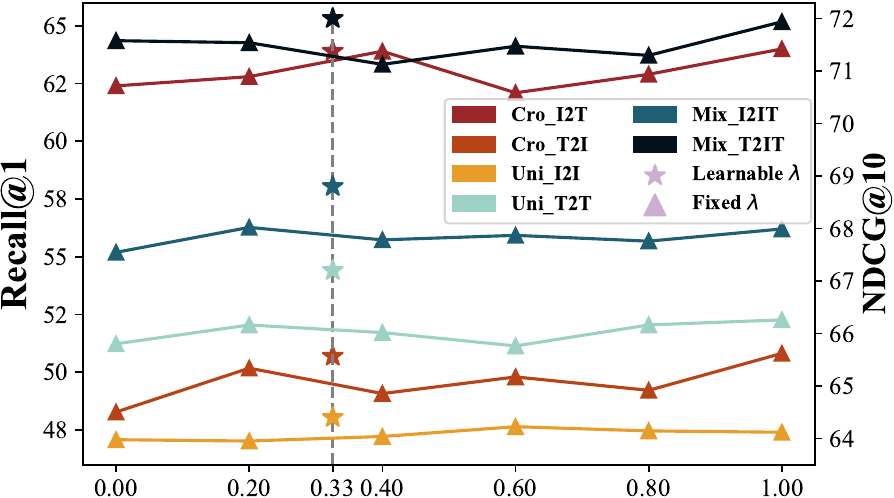}\\
\mbox{  ({\it f}) $\lambda$ analysis (Vizwiz).}
\end{minipage}
\caption{\textbf{Parameter analyses}. We verify the influence of parameters our method under FLICKR30K and Vizwiz datasets.}\label{fig:f4}
}
\end{figure*}

\begin{table*}[t]{\small
    \centering
    \caption{Compare MAE with MSE, KL, and WD. Evaluation criteria are R$@$A and NDCG$@$A.}
    \label{tab:tab5}
    \begin{tabular*}{0.985\textwidth}{@{\extracolsep{\fill}}@{}l@{\hspace{0.06cm}}|@{}c@{\hspace{0.06cm}}|@{}c@{\hspace{0.06cm}}|@{}c@{\hspace{0.06cm}}|@{}c@{\hspace{0.06cm}}|@{}c@{\hspace{0.06cm}}|@{}c@{\hspace{0.06cm}}|@{}c@{\hspace{0.06cm}}|@{}c@{\hspace{0.06cm}}|@{}c@{\hspace{0.06cm}}|@{}c@{\hspace{0.06cm}}|@{}c@{\hspace{0.06cm}}|@{}c@{\hspace{0.06cm}}}
    \hline
    \multirow{3}{*}{Methods} &\multicolumn{6}{c|}{FLICKR30K} & \multicolumn{6}{c}{Vizwiz}\\
    \cline{2-13}
    & \multicolumn{3}{c|}{I2T} & \multicolumn{3}{c|}{T2I} & \multicolumn{3}{c|}{I2T} & \multicolumn{3}{c}{T2I}\\
    \cline{2-13}
    & $@$1 & $@$5 & $@$10 & $@$1 & $@$5 & $@$10 & $@$1 & $@$5 & $@$10 & $@$1 & $@$5 & $@$10\\
    \hline
    X-VLM*+ (MAE) &\underline{85.6} &\textbf{98.1} &\textbf{99.3} &\textbf{73.3} &\textbf{93.0} &\textbf{96.1} &\textbf{63.9} &\textbf{85.7} &\textbf{90.6} &\textbf{50.7} &\underline{75.5} &\textbf{83.7}\\
    X-VLM*+ (MSE) &84.0 &\underline{97.6} &\underline{99.0} &69.6 &91.7 &95.1 &60.9 &84.4 &\textbf{90.6} &\underline{48.5} &\textbf{75.7} &\underline{83.4} \\
    X-VLM*+ (KL) &7.3 &22.8 &36.3 &1.8 &7.1 &11.7 &8.9 &25.3 &\underline{38.5} &4.1 &14.2 &23.4 \\
    X-VLM*+ (WD) &\textbf{85.8} &97.5 &\underline{99.0} &\underline{70.7} &\underline{92.1} &\underline{95.2} &\underline{63.1} &\underline{85.1} &\textbf{90.6} &{48.4} &75.4 &\textbf{83.7} \\
    \hline
    & \multicolumn{3}{c|}{I2I} & \multicolumn{3}{c|}{T2T} & \multicolumn{3}{c|}{I2I} & \multicolumn{3}{c}{T2T}\\
    \cline{2-13}
    & $@$10 & $@$20 & $@$50 & $@$10 & $@$20 & $@$50 & $@$10 & $@$20 & $@$50 & $@$10 & $@$20 & $@$50\\
    \hline
    X-VLM*+ (MAE) &\textbf{64.3} &\textbf{65.4} &\underline{67.8} &\textbf{73.0} &\textbf{71.4} &\textbf{68.9} &\textbf{64.4} &\textbf{66.1} &\textbf{68.9} &\textbf{67.2} &\textbf{66.5} &{65.5}\\
    X-VLM*+ (MSE) &63.8 &\underline{65.2} &\textbf{68.0} &\textbf{73.0} &\underline{70.9} &68.1 &64.1 &\underline{65.9} &\underline{68.8} &\underline{65.8} &\underline{65.8} &\underline{65.6} \\
    X-VLM*+ (KL) &61.5 &63.1 &66.1 &71.5 &70.0 &68.0 &63.2 &64.9 &68.1 &61.6 &62.5 &63.2 \\
    X-VLM*+ (WD) &\underline{64.0} &65.1 &\underline{67.8} &\underline{72.8} &70.8 &\underline{68.2} &\underline{64.3} &\underline{65.9} &\underline{68.8} &\underline{65.8} &\underline{65.8} &\textbf{65.7} \\
    \hline
    & \multicolumn{3}{c|}{I2IT} & \multicolumn{3}{c|}{T2IT} & \multicolumn{3}{c|}{I2IT} & \multicolumn{3}{c}{T2IT}\\
    \cline{2-13}
    & $@$10 & $@$20 & $@$50 & $@$10 & $@$20 & $@$50 & $@$10 & $@$20 & $@$50 & $@$10 & $@$20 & $@$50\\
    \hline
    X-VLM*+ (MAE) &\textbf{79.1} &\underline{74.0} &\textbf{68.8} &\textbf{79.6} &\textbf{76.7} &\underline{72.5} &\textbf{68.8} &\textbf{66.3} &\underline{63.1} &\textbf{72.0} &\textbf{70.2} &\textbf{67.6} \\
    X-VLM*+ (MSE) &\underline{78.2} &73.2 &68.3 &78.4 &\underline{76.4} &72.1 &66.9 &66.1 &\textbf{63.7} &67.5 &67.5 &66.6  \\
    X-VLM*+ (KL) &48.4 &48.8 &50.3 &75.1 &73.5 &71.4 &51.3 &51.8 &53.5 &67.6 &66.7 &65.5\\
    X-VLM*+ (WD) &78.0 &\textbf{74.5} &\underline{68.6} &\underline{78.5} &75.6 &\textbf{73.2} &\underline{67.1} &\underline{66.2} &\textbf{63.7} &\underline{68.0} &\underline{67.9} &\underline{66.8} \\
    \hline
    \end{tabular*}
}
\end{table*}

\noindent{\bf{Single-Modal Retrieval:}} Conducting two tasks, i.e., I2I and T2T, we aim to verify whether traditional vision-language models can preserve structural information of single modality after cross-modal representation learning. 
% Figture~\ref{fig:single_mixed} (a--d) exhibits the single-modal retrieval performance between traditional models and vision-language pre-training models.
% \blue{Figture~\ref{fig:single_mixed} (a--d) and (e--h) represent the results of I2I NDCG$@$10 and T2T NDCG$@$10, respectively.} 
Table \ref{tab:tab2} exhibits the single-modal retrieval performance between traditional models and vision-language pre-training models.
Note that ``SWIN'' and ``BERT'' represent directly utilizing unsupervised prototype-aware contrastive learning for training, and then test the single-modal retrieval performance. Experiment results indicate that: 1). Our X-VLM*+ performs better than nearly all cross-modal retrieval comparison methods in both I2I and T2T retrievals on four datasets. It even outperforms the best single-modal retrieval methods, on I2I NDCG$@$10 and T2T NDCG$@$10 by 3.6/4.8/0.2/14.1 and 2.4/2.4/1.4/11.7 respectively. 2). Although several cross-modal retrieval methods can improve the I2I retrieval, e.g., ALBEF* increases the NDCG@10 compared with Swin Transformer (1.7/2.8 on MS-COCO (1K)/MS-COCO (5K)), almost all methods exhibit varying degrees of performance decline in T2T retrieval, with the improvement in the weak modality's performance being less pronounced than the decline in the strong modality's performance. For instance, ALBEF* decreases 21.5/15.9 of T2T NDCG@10 on MS-COCO (1K) and MS-COCO (5K) datasets compared with BERT, with only 1.7/2.8 promotion of I2I NDCG@10 over Swin Transformer. 3). Our X-VLM*+ performs better than CYCLIP*, which reveals that adaptively distillation is superior to learning structure-preserving representations by better corporating structure consistency.

\noindent{\bf{Mixed Retrieval:}} It further simulates the real task that retrieves all modal instances with single-modal query (i.e., image or text). Figures~\ref{fig:single_mixed} (a--d) and (e--h) represent the results of I2IT NDCG$@$10 and T2IT NDCG$@$10, respectively, which reveal that: 1). As the position of the expected instances $A$ increases, the performance of several models degrades. Since the retrieval library has limited capacity, even though most expected instances rank high, the remaining results may have low similarity, causing a decline in the similarity-based metric NDCG. 2). Our X-VLM*+ outperforms nearly all comparison methods on four datasets, which validates that our method can effectively learn cross-modal consistent and structure-preserving representations simultaneously. More comprehensive experimental results can be found in Section~\uppercase\expandafter{\romannumeral2} of the supplementary due to space limitation.

% 4.9 pre trained
\begin{table*}[h!]{\small
\centering
\caption{Performance of cross-modal, single-modal, and mix-modal retrieval with pre-trained models on FLICKR30K dataset. Evaluation criteria are R$@$A and NDCG$@$A. The method with “+" sign is our method.}
\label{tab:tab7}

\begin{tabular*}{0.985\textwidth}{@{\extracolsep{\fill}}@{}c@{\hspace{0.1cm}}|@{}c@{\hspace{0.1cm}}|@{}c@{\hspace{0.1cm}}|@{}c@{\hspace{0.1cm}}|@{}c@{\hspace{0.1cm}}|@{}c@{\hspace{0.1cm}}|@{}c@{\hspace{0.1cm}}|@{}c@{\hspace{0.1cm}}|@{}c@{\hspace{0.1cm}}|@{}c@{\hspace{0.1cm}}|@{}c@{\hspace{0.1cm}}|@{}c@{\hspace{0.1cm}}|@{}c@{\hspace{0.1cm}}|@{}c@{\hspace{0.1cm}}|@{}c@{\hspace{0.1cm}}|@{}c@{\hspace{0.1cm}}|@{}c@{\hspace{0.1cm}}|@{}c@{\hspace{0.1cm}}|@{}c@{\hspace{0.1cm}}}
\hline
\multirow{3}{*}{Methods} & \multicolumn{6}{c|}{Cross-Modal Retrieval} & \multicolumn{6}{c|}{Single-Modal Retrieval} & \multicolumn{6}{c}{Mixed Retrieval}\\
\cline{2-19}
& \multicolumn{3}{c|}{I2T} & \multicolumn{3}{c|}{T2I} 	& \multicolumn{3}{c|}{I2I} & \multicolumn{3}{c|}{T2T} & \multicolumn{3}{c|}{I2IT} & \multicolumn{3}{c}{T2IT}\\
\cline{2-19}
& $@$1 & $@$5 & $@$10 & $@$1 & $@$5 & $@$10  & $@$10 & $@$20 & $@$50 & $@$10 & $@$20 & $@$50 & $@$10 & $@$20 & $@$50 & $@$10 & $@$20 & $@$50\\
\hline
X-VLM &95.5  &99.8 &\textbf{100.0} &85.3  &97.0  &98.5 & 64.1 & 65.6 & 68.3 & 74.9 & 72.8 & 70.3 & 85.5 & 79.2 & 71.4 & 82.0 & 78.6 & 74.1 \\	
X-VLM+ &\textbf{96.6} &\textbf{100.0} &\textbf{100.0} &\textbf{86.8} &\textbf{97.7} &\textbf{99.0} &\textbf{64.3} &\textbf{65.7} &\textbf{68.5} &\textbf{75.3} &\textbf{73.4} &\textbf{70.9} &\textbf{86.5} &\textbf{79.6} &\textbf{71.6} &\textbf{82.4} &\textbf{79.2} &\textbf{74.7} \\
\hline
\end{tabular*}}
\end{table*}

\begin{figure*}[!htb]\centering
\centering
\includegraphics[width = 170mm]{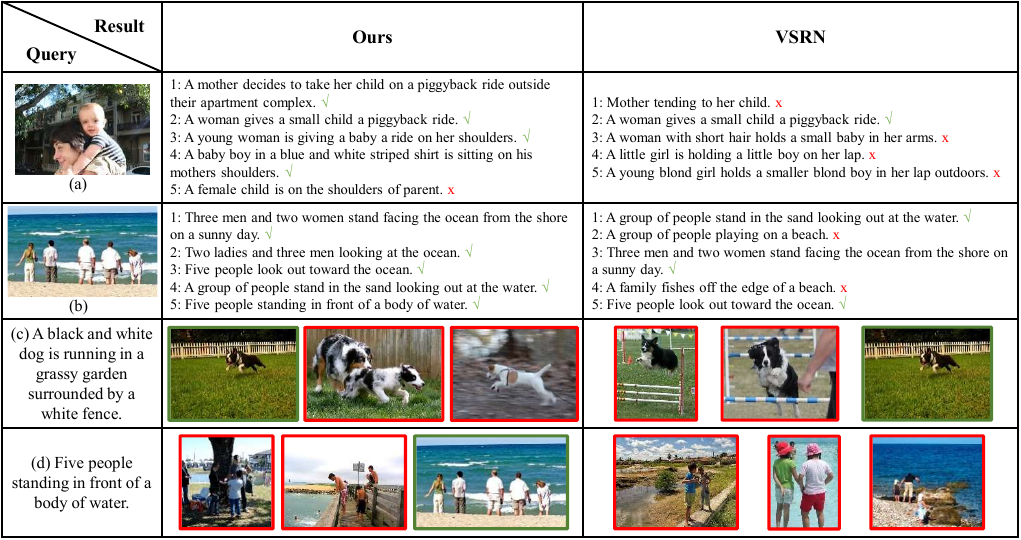}\\
\caption{\textbf{Qualitative results of cross-modal retrieval}. For each image query, we show the top-5 ranked sentences. For each sentence query, we show the top-3 ranked images, ranking from left to right (Best viewed in green). The examples are sampled from the FLICKR30K dataset.}\label{fig:f2}
\end{figure*}

\subsection{Ablation Study}
Furthermore, we conduct ablation studies to validate the effectiveness of each module: cross-modal matching ($\ell_{itc}$ and $\ell_{itm}$), representation-level distillation losses ($\ell_{iic}$ and $\ell_{ttc}$) and the structure-aware distillation module ($\ell_{sa}$). Due to the large size of the MS-COCO dataset, we focus on the smaller FLICKR30K and Vizwiz datasets in subsequent experiments. Analyzing the results in Table~\ref{tab:tab4} reveals the following observations: 1). The removal of $\ell_{itm}$ leads to a significant decline in cross-modal retrieval performance, although single-modal retrieval performance improves, further confirming the negative impact of cross-modal consistency learning on single-modal retrieval. 2). Compared to the removal of $\ell_{itc}$ and $\ell_{itm}$, the elimination of $\ell_{sa}$ results in the poorest retrieval performance, indicating that preserving instance structure information more effectively promotes representation learning both across modalities and within each modality.

% Furthermore, we conduct ablation studies to validate the effectiveness of each multi-granularity distillation module: the representation-level distillation loss ($\ell_{iic}$ and $\ell_{ttc}$) and the structure-aware distillation module ($\ell_{sa}$). Due to the large size of the MS-COCO dataset, we focus on the smaller FLICKR30K and Vizwiz datasets in subsequent experiments. The results are documented in Table \ref{tab:tab4} and reveal the following observations: 1). The removal of $\ell_{sa}$ leads to the poorest retrieval performance, indicating that preserving instance structure information effectively promotes representation learning both across modalities and within each modality. 2). The fusion of all modules yields the optimal performance, suggesting that each module contributes to enhancing overall performance.

\subsection{Sensitivity to Parameters}
To verify the influence of parameters, we conduct more experiments by tuning several important parameters: 1). batch size $J$; 2). temperature parameter $\tau$; 3). hyper-parameter $\lambda$.

\noindent {\bf{Influence of Batch Size:}} We incorporate batch with different sizes, i.e., $\{12, 24, 36, 48\}$ into the proposed model, to empirically investigate the impact of batch size on performance. The performance in Fig.~\ref{fig:f4} (a) and (d) first increases and then decreases, indicating that a larger batch size can consider more neighbor information, but an oversized one may introduce noisy information.

\noindent {\bf{Influence of Temperature Parameter $\tau$:}} To explore the influence of the temperature scale parameter, we tune the $\tau \in \{0.1, 0.2, 0.5, 2, 3\}$ to conduct more experiments. Fig.~\ref{fig:f4} (b) and (e) record the results. We find that the retrieval results are the best when $\tau=0.1$ on two datasets. This indicates that the points have few similar neighbors, which can promote the learning of structure-aware representations. 

\noindent {\bf{Influence of Hyper-Parameter $\lambda$:}} To investigate the importance of vision and language modalities in Equations \ref{eq:e2} and \ref{eq:e10}, Figures \ref{fig:f4} (c) and (f) present the results under fixed $\lambda \in \{0, 0.2, 0.4, 0.6, 0.8, 1.0\}$ and learnable $\lambda$. We have observed the following phenomenons: 1). All five-star markers are positioned above the triangles on the same-colored lines, which indicates that the learnable $\lambda$ effectively contributes to the discovery of the most optimal modality fusion for the model. 2). Specifically, on both the FLICKR30K and Vizwiz datasets, the ultimate values of learnable $\lambda$ are $0.21$ and $0.33$, respectively. This indicates that the language modality (i.e., the strong modality) plays a more pivotal role in information provision during the distillation process.

\begin{figure*}[!htb]\centering
\centering
\includegraphics[width = 170mm]{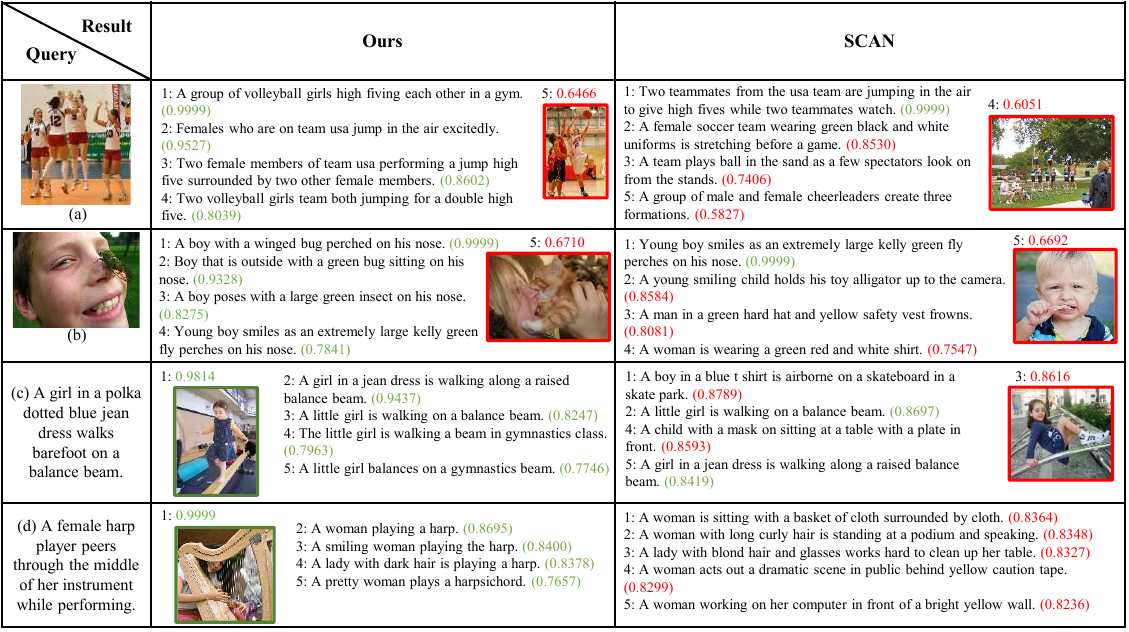}\\
\caption{\textbf{Qualitative results of mixed retrieval}. For each query, we show the top-5 ranked instances, including images and sentences (Correct results viewed in green). The examples are sampled from the FLICKR30K dataset.}\label{fig:f5}
\end{figure*}

\subsection{Influence with Different Distillation Measurements}
To explore the effectiveness of graph-matching criteria, we conducted more experiments. To be specific, we replace the MAE distance with other distance measurements, i.e., the Mean Squared Error (MSE), Kullback-Leibler divergence (KL), and Wasserstein distance (WD). The results in Table \ref{tab:tab5} reveal that our graph-matching criteria, independent of specific distance metrics, demonstrate excellent flexibility, as observed in the MAE, MSE, and WD outcomes. In contrast, the KL method is not conducive to cross-modal retrieval.

\begin{table}[htbp]{
\centering
\caption{Performance of the multi-granularity distillation strategy on different retrieval models (marked with ``+''). \\Evaluation criteria are R$@$A and NDCG$@$A.}
\label{tab:tab6}
\begin{tabular*}{0.49\textwidth}{@{\extracolsep{\fill}}@{}l@{\hspace{0.06cm}}|@{}c@{\hspace{0.06cm}}|@{}c@{\hspace{0.06cm}}|@{}c@{\hspace{0.06cm}}|@{}c@{\hspace{0.06cm}}|@{}c@{\hspace{0.06cm}}|@{}c@{\hspace{0.06cm}}|@{}c@{\hspace{0.06cm}}|@{}c@{\hspace{0.06cm}}|@{}c@{\hspace{0.06cm}}|@{}c@{\hspace{0.06cm}}|@{}c@{\hspace{0.06cm}}|@{}c@{\hspace{0.06cm}}}
\hline
\multirow{3}{*}{Methods} &\multicolumn{6}{c|}{FLICKR30K} & \multicolumn{6}{c}{Vizwiz}\\
\cline{2-13}
& \multicolumn{3}{c|}{I2T} & \multicolumn{3}{c|}{T2I} & \multicolumn{3}{c|}{I2T} & \multicolumn{3}{c}{T2I}\\
\cline{2-13}
& $@$1 & $@$5 & $@$10 & $@$1 & $@$5 & $@$10 & $@$1 & $@$5 & $@$10 & $@$1 & $@$5 & $@$10\\
\hline
SCAN*  &60.0 &83.9 &90.7 &37.7 &66.3 &76.0 &34.9 &60.9 &72.8 &24.8 &48.6 &59.0 \\
SCAN*+ &\textbf{61.0} &\textbf{86.5} &\textbf{92.5} &\textbf{40.8} &\textbf{72.3} &\textbf{81.2} &\textbf{38.3} &\textbf{64.4} &\textbf{73.4} &\textbf{27.6} &\textbf{52.9} &\textbf{62.8} \\
VSRN* &58.0 &86.1 &91.6 &46.9 &77.0 &85.1 &34.8 &63.1 &73.5 &26.3 &52.5 &64.1 \\
VSRN*+ &\textbf{62.1} &\textbf{86.3} &\textbf{92.1} &\textbf{47.4} &\textbf{77.2} &\textbf{85.3} &\textbf{35.9} &\textbf{64.3} &\textbf{73.7} &\textbf{28.1} &\textbf{55.1} &\textbf{67.4} \\
ALBEF* &63.2 &87.4 &93.5 &48.5 &73.1 & 80.7 &47.7 & 70.4 &79.6 &34.3 &56.3 &65.7\\
ALBEF*+ &\textbf{65.8} &\textbf{88.6} &\textbf{94.8} &\textbf{50.7} &\textbf{75.3} &\textbf{82.6} &\textbf{50.2} &\textbf{73.3} &\textbf{82.0} &\textbf{36.9} &\textbf{61.2} &\textbf{70.1} \\
\hline
& \multicolumn{3}{c|}{I2I} & \multicolumn{3}{c|}{T2T} & \multicolumn{3}{c|}{I2I} & \multicolumn{3}{c}{T2T}\\
\cline{2-13}
& $@$10 & $@$20 & $@$50 & $@$10 & $@$20 & $@$50 & $@$10 & $@$20 & $@$50 & $@$10 & $@$20 & $@$50\\
\hline
SCAN* &52.6 &55.0 &59.1 & 38.7 &42.6 &47.6 & 55.4 &57.4 &61.1 &52.0 &54.0 & 56.5 \\
SCAN*+  &\textbf{53.4} &\textbf{55.6} &\textbf{59.6} &\textbf{45.6} &\textbf{49.5} &\textbf{54.3} &\textbf{58.5} &\textbf{60.5} &\textbf{64.0} &\textbf{53.4} &\textbf{56.8} &\textbf{60.4} \\
VSRN* & 61.7 & 63.4 & 66.4 & {67.8} & {67.4} & {66.3} & 62.0 &63.7 &66.7 & 59.8 &60.8 &61.6\\
VSRN*+ &\textbf{63.5} &\textbf{64.9} &\textbf{67.3} &\textbf{68.0} &\textbf{67.8} &\textbf{66.8} &\textbf{62.6} &\textbf{64.4} &\textbf{67.5} &\textbf{60.2} &\textbf{61.4} &\textbf{62.4} \\
ALBEF* &\textbf{60.8} &62.3 &\textbf{65.2} &70.0 &68.4 & 66.2 &\textbf{61.8} & 63.3 &66.2 &63.7 &63.0 &62.1\\
ALBEF*+  &\textbf{60.8} &\textbf{62.4} &\textbf{65.2} &\textbf{72.2} &\textbf{70.3} &\textbf{68.1} &\textbf{61.8} &\textbf{63.4} &\textbf{66.4} &\textbf{65.3} &\textbf{64.5} &\textbf{63.4} \\
\hline
& \multicolumn{3}{c|}{I2IT} & \multicolumn{3}{c|}{T2IT} & \multicolumn{3}{c|}{I2IT} & \multicolumn{3}{c}{T2IT}\\
\cline{2-13}
& $@$10 & $@$20 & $@$50 & $@$10 & $@$20 & $@$50 & $@$10 & $@$20 & $@$50 & $@$10 & $@$20 & $@$50\\
\hline
SCAN* &63.2 &60.8 &59.2 &63.2 &60.8 &59.3 &64.9 &62.3 &60.0 &64.9 &62.4 &60.2\\
SCAN*+  &\textbf{67.1} &\textbf{65.5} &\textbf{64.4} &\textbf{67.1} &\textbf{65.6} &\textbf{64.5} &\textbf{65.8} &\textbf{64.7} &\textbf{62.4} &\textbf{71.4} &\textbf{70.0} &\textbf{65.5} \\
VSRN* & 73.8 & 71.6 & 68.5 & {73.9} & {71.7} & {68.7} &65.4 &63.8 &{63.4} &65.7 &64.2 &63.7\\
VSRN*+ &\textbf{77.4} &\textbf{73.4} &\textbf{69.3} &\textbf{79.5} &\textbf{76.5} &\textbf{72.5} &\textbf{69.0} &\textbf{65.9} &\textbf{65.1} &\textbf{72.0} &\textbf{70.1} &\textbf{66.6} \\
ALBEF* &48.1 &48.3 &\textbf{49.7} &73.8 &71.1 & 67.9 &49.3 &49.3 &50.5 &69.4 &67.0 &64.3\\
ALBEF*+  &\textbf{48.2} &\textbf{48.4} &\textbf{49.7} &\textbf{75.3} &\textbf{72.7} &\textbf{69.5} &\textbf{59.2} &\textbf{57.3} &\textbf{56.0} &\textbf{70.6} &\textbf{68.3} &\textbf{65.5} \\
\hline
\end{tabular*}}
\end{table}
\subsection{Generalization of Multi-Granularity Distillation}
To assess the generalization ability of our multi-granularity distillation strategy, we extended experiments to include the retrieval methods SCAN and VSRN based on dual encoders, along with the transformer-based retrieval method ALBEF. These experiments were conducted on the FLICKR30K and Vizwiz datasets. SCAN adopts a local representation matching approach, while VSRN introduces a global semantic reasoning module alongside attention to region relationships. As dual encoder methods do not learn modal fusion representations, we introduced the structure-aware distillation module for each single modality, aiming to enhance instance-level modality matching by leveraging the representation structure information from the single-modal teacher model. Results in Table~\ref{tab:tab6} show a significant performance improvement when our multi-granularity distillation strategy (denoted as ``+'') is combined with various methods. This phenomenon validates the effectiveness and generalization capability of our approach.

\subsection{Exploration on Large-Scale Multi-modal Model}
To explore the effectiveness of our proposed coordinate optimization strategy for large-scale multi-modal models, we apply our method to the fine-tuning process (marked with ``+''). Table \ref{tab:tab7} takes X-VLM as an example on the FLICKR30K dataset. Results demonstrate that our method not only improves the performance of single-modal retrieval in the fine-tuning stage of the pre-trained large-scale model, for both the weak and strong modalities but also maintains excellent performance on most metrics in cross-modal retrieval and mixed retrieval tasks.

\subsection{Case Study} 
To analyze the retrieval visualizations, we randomly sample cross-modal and mixed retrieval cases from the FLICKR30K dataset to validate the effectiveness of our proposed method. The visualization examples are exhibited in Fig.~\ref{fig:f2} and Fig.~\ref{fig:f5}, in which green ticks/boxes/values represent exactly aligned instances, red forks/boxes/values denote unaligned instances, the mixed retrieval instances are with ROUGE-L value (the larger the better). Considering the superior performance, we adopt our method here.

Fig.~\ref{fig:f2} shows the qualitative results of cross-modal retrieval using our method and VSRN. First, most of the retrieved cross-modal instances using our method are correct (shown as green ticks) on both the I2T and T2I retrieval. Some outputs are mismatched (shown as red forks), but reasonable, for example, (a) 5 contains similar semantic meanings to the image. On the other hand, our method outperforms the VSRN on both the I2T and T2I retrieval considering the same query. For example, (b) and (d) share the same image query, our method can retrieve the most aligned sentences, while VSRN fails, the reason that better structure-preserving can promote consistent representation learning in return.

Fig.~\ref{fig:f5} shows the qualitative results of mixed retrieval using our method compared with the SCAN. We find that our method can not only find accurate cross-modal instances but also find semantically similar intra-modal instances. For example, in the image query case (a), our method can retrieve similar images and exactly aligned cross-modal instances, while SCAN only retrieves images with lower similarities (i.e., lower ROUGE-L values), and several unaligned sentences. This phenomenon further validates the effectiveness of our proposed method in mixed retrieval. Moreover, we can intuitively find the differences between consistent representations learned by different cross-modal approaches in (d). Using the ``harp player'' case as an example, we can find the clustered instances in our method are all with ``harp player'' semantics, whereas the clustered instances in SCAN are with outliers.
\section{Conclusion}
In this paper, we find that the learned consistent representations from existing vision-language retrieval methods may affect single-modal retrieval performance. To explain this phenomenon, we identify the main cause as modal sufficiency, i.e., there exist weak and strong modalities, and hard cross-modal consistency may bring negative representation learning to strong modality, leading to the destruction of instance structure. To address this problem, we develop a different way inspired by multi-task learning, which aims to learn two modal representations that can simultaneously ensure cross-modal consistency and single-modal structure. Extensive experiments on different datasets validate our method can achieve better single-modal retrieval accuracy whilst maintaining cross-modal retrieval capacity compared with the baselines. 
In future work, we aspire to theoretically elucidate the relationship between the degree of modality imbalance and retrieval performance. Additionally, addressing the challenges posed by modality imbalance in cross-modal retrieval tasks involving more than two modalities represents a significant area for further investigation.

\normalem
\bibliography{IEEEabrv,paper.bib}
\bibliographystyle{IEEEtran}

\begin{IEEEbiography}[{\includegraphics[width=1in,height=1.25in,clip,keepaspectratio]{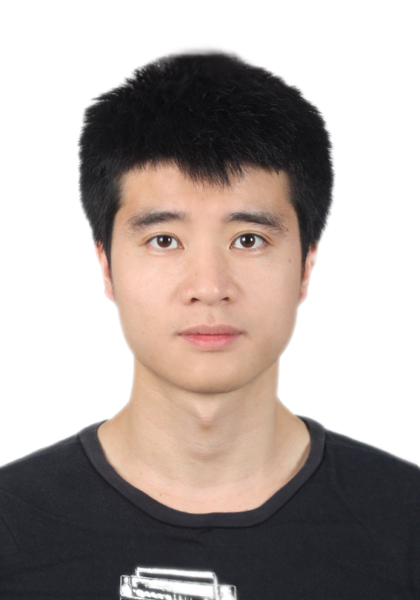}}]{Yang Yang} (Member, IEEE) received a Ph.D. degree in computer science, from Nanjing University, China in 2019. In the same year, he became a faculty member at Nanjing University of Science and Technology, China. He is currently a Professor at the School of Computer Science and Engineering. His research interests lie primarily in machine learning and data mining, including heterogeneous learning, model reuse, and incremental mining. He has published over 20 papers in leading international journals/conferences. He serves as PC in leading conferences such as IJCAI, AAAI, ICML, NIPS, etc.
\end{IEEEbiography}
\begin{IEEEbiography}[{\includegraphics[width=1in,height=1.25in,clip,keepaspectratio]{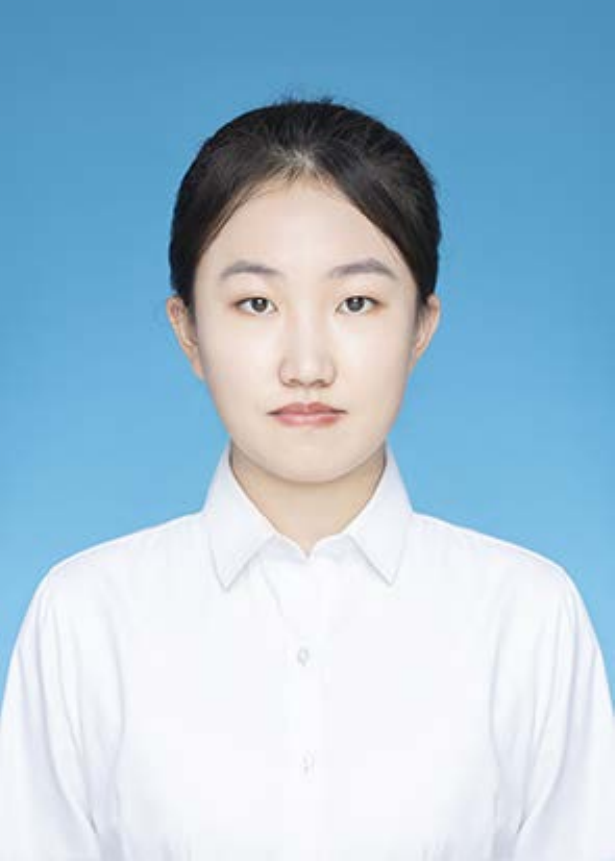}}] {Wenjuan Xi} (Student Member, IEEE) is working towards the M.Sc. degree from the School of Computer Science and Engineering, at Nanjing University of Science and Technology, China. Her research interests lie primarily in machine learning and data mining, including multi-modal learning.
\end{IEEEbiography}
\begin{IEEEbiography}[{\includegraphics[width=0.9in,height=1.25in,clip,keepaspectratio]{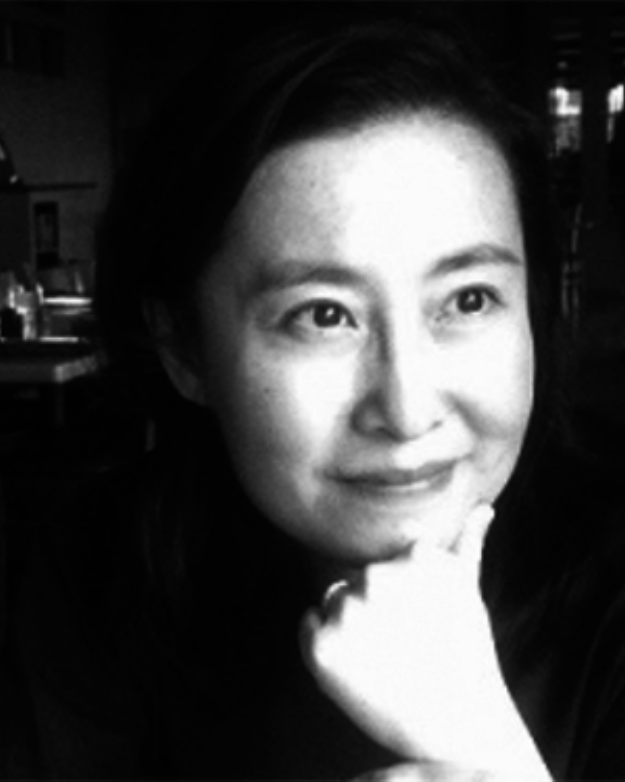}}] {Luping Zhou} (Senior Member, IEEE) received the PhD degree from Australian National University, Canberra, Australia. She is now a senior lecturer with the School of Electrical and Information Engineering, University of Sydney, Australia. She was a recipient of Australian Research Council DECRA award (Discovery Early Career Researcher Award) in 2015. Her research interests include machine learning, computer vision, and medical image analysis.
\end{IEEEbiography}
\begin{IEEEbiography}[{\includegraphics[width=1in,height=1.25in,clip,keepaspectratio]{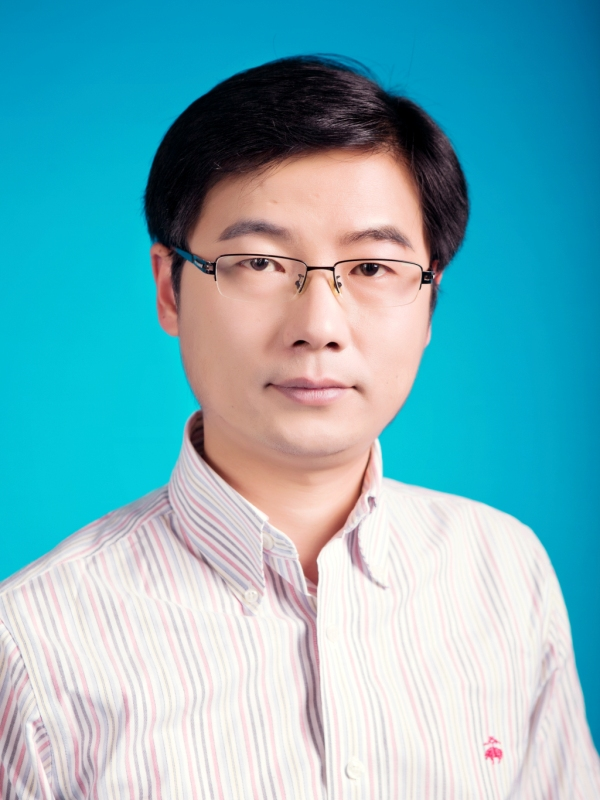}}]{Jinhui Tang} (Senior Member, IEEE) received the B.E. and Ph.D. degrees from the University of Science and Technology of China, Hefei, China, in 2003 and 2008, respectively. He is currently a Professor with the Nanjing University of Science and Technology, Nanjing, China. He has authored more than 200 articles in top-tier journals and conferences. His research interests include multimedia analysis and computer vision. Dr. Tang was a recipient of the Best Paper Awards in ACM MM 2007, PCM 2011, ICIMCS 2011, and ACM MM Asia 2020, the Best Paper Runner-Up in ACM MM 2015, and the Best Student Paper Awards in MMM 2016 and ICIMCS 2017. He has served as an Associate Editor for the IEEE TMM, IEEE TNNLS, the IEEE TKDE, and the IEEE TCSVT. He is a Fellow of IAPR.
\end{IEEEbiography}
\vfill

\clearpage

\end{document}